\documentclass[a4paper,fleqn]{cas-dc}

\usepackage[numbers]{natbib}

\usepackage{subfigure}

\usepackage{xspace}

\usepackage{algorithm}
\usepackage{algorithmic}
\usepackage{caption2}
\makeatletter
\DeclareRobustCommand\onedot{\futurelet\@let@token\@onedot}
\def\@onedot{\ifx\@let@token.\else.\null\fi\xspace}
\def\eg{\emph{e.g}\onedot} 
\def\ie{\emph{i.e}\onedot}

\def\etal{\emph{et al}\onedot}
\makeatother

\def\tsc#1{\csdef{#1}{\textsc{\lowercase{#1}}\xspace}}
\tsc{WGM}
\tsc{QE}
\tsc{EP}
\tsc{PMS}
\tsc{BEC}
\tsc{DE}
\begin{document}
\let\WriteBookmarks\relax
\def\floatpagepagefraction{1}
\def\textpagefraction{.001}

\shorttitle{ Z. Liu and L. Zhu / Neurocomputing}
\shortauthors{Zhikang Liu et~al.}

\title [mode = title]{Label-guided Attention Distillation for Lane Segmentation}

\author[1]{Zhikang Liu}[style=chinese,
						type=editor,
                        orcid=0000-0002-5106-697X]
\cormark[1]
\ead{lzk@mail.ustc.edu.cn}
%\credit{Conceptualization, Methodology, Writing - original draft, Writing - review \& editing}

\address[1]{Megvii Technology Limited, Beijing 100086, China}

\author[2]{Lanyun Zhu}[style=chinese]
\ead{zhulanyun@buaa.edu.cn}

% \credit{Data curation, Software, Validation, Writing - original draft}

\address[2]{School of Instrumentation and Optoelectronic Engineering, Beihang University, Beijing 100191, China}

\cortext[cor1]{Corresponding author}

\begin{abstract}
Contemporary segmentation methods are usually based on deep fully convolutional networks (FCNs). However, the layer-by-layer convolutions with a growing receptive field is not good at capturing long-range contexts such as lane markers in the scene. In this paper, we address this issue by designing a distillation method that exploits label structure when training segmentation network. The intuition is that the ground-truth lane annotations themselves exhibit internal structure. We broadcast the structure hints throughout a teacher network, i.e., we train a teacher network that consumes a lane label map as input and attempts to replicate it as output. Then, the attention maps of the teacher network are adopted as supervisors of the student segmentation network. The teacher network, with label structure information embedded, knows distinctly where the convolution layers should pay visual attention into. The proposed method is named as Label-guided Attention Distillation (LGAD). It turns out that the student network learns significantly better with LGAD than when learning alone. As the teacher network is deprecated after training, our method do not increase the inference time. Note that LGAD can be easily incorporated in any lane segmentation network.

To validate the effectiveness of the proposed LGAD method, extensive experiments have been conducted on two popular lane detection benchmarks: TuSimple and CULane. The results show consistent improvement across a variety of convolutional neural network architectures. Specifically, we demonstrate the accuracy boost of LGAD on the lightweight model ENet. It turns out that the ENet-LGAD surpasses existing lane segmentation algorithms.

\end{abstract}


\maketitle

\section{Introduction}

Lane segmentation has attracted increasing attention recently, owing to its applications in driving assistance systems and self-driving vehicles. This visual task aims at locating lanes in road scene images. The precise locations of lanes are beneficial to downstream tasks such as positioning cars within lanes, lanes departure detection, and trajectory planning.

\begin{figure}[t]
	\centering
	\centerline{\includegraphics[width=8cm]{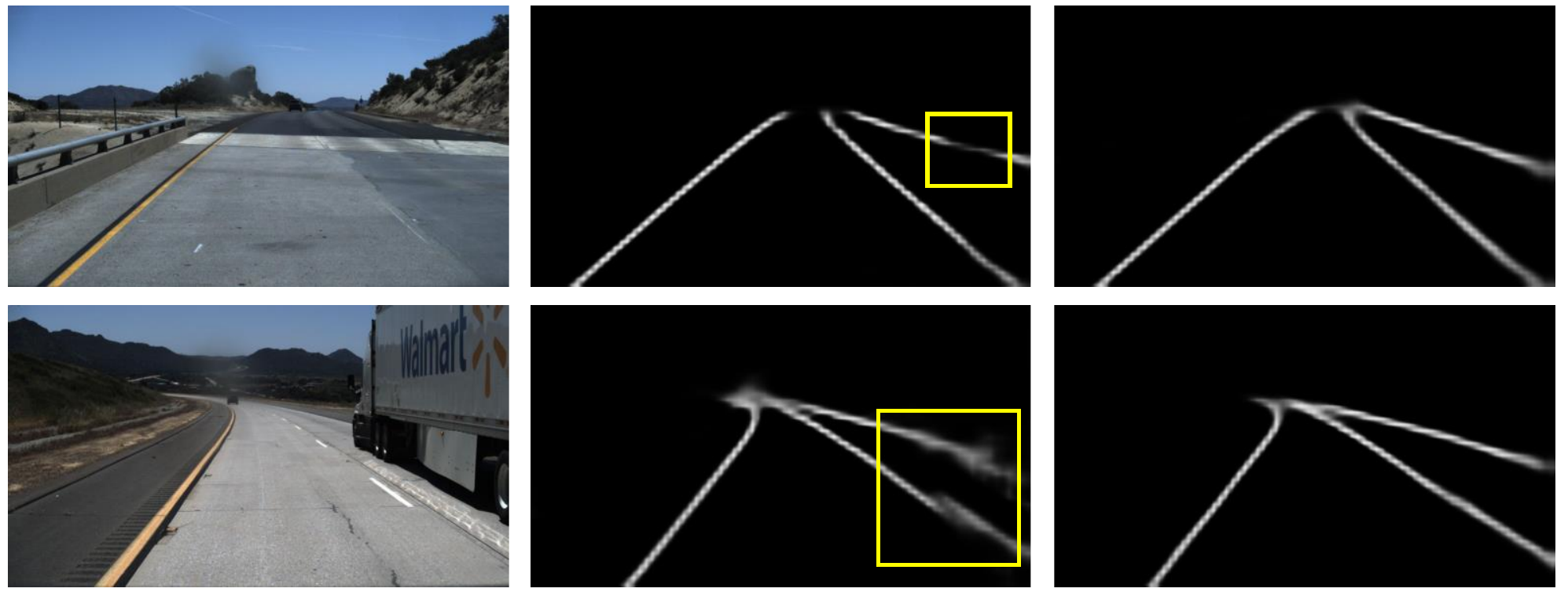}}
	\caption{Comparison between FCN and FCN-LGAD in lane segmentation. For each example, from left to right are input image, output of FCN, output of FCN-LGAD. It can be seen that FCN suffers from blur on the boundaries and disconnection caused by occlusions. FCN-LGAD could precisely capture the long continuous shape of lane markings and fix the disconnected parts in FCN. (best viewed in color)}
	\label{fig_first}
\end{figure}

Contemporary lane segmentation methods \cite{ghafoorian2018gan,hou2019learning} are typically based on deep learning using fully convolutional networks (FCNs) and aim to assign a binary label to each pixel in an image to indicate whether it belongs to a lane mark or not. An inherent challenge for lane segmentation is that CNN based segmentation methods still not perform well for long-range narrow-width objects, as the linearly growing receptive field is not good at precisely capturing long continuous shape contexts. As illustrated in red bounding boxes in Figure~\ref{fig_first}, traditional FCNs suffer from blur on the lane boundaries and disconnection caused by occlusions.

\begin{figure}[t]
	\centering
	\centerline{\includegraphics[width=8cm]{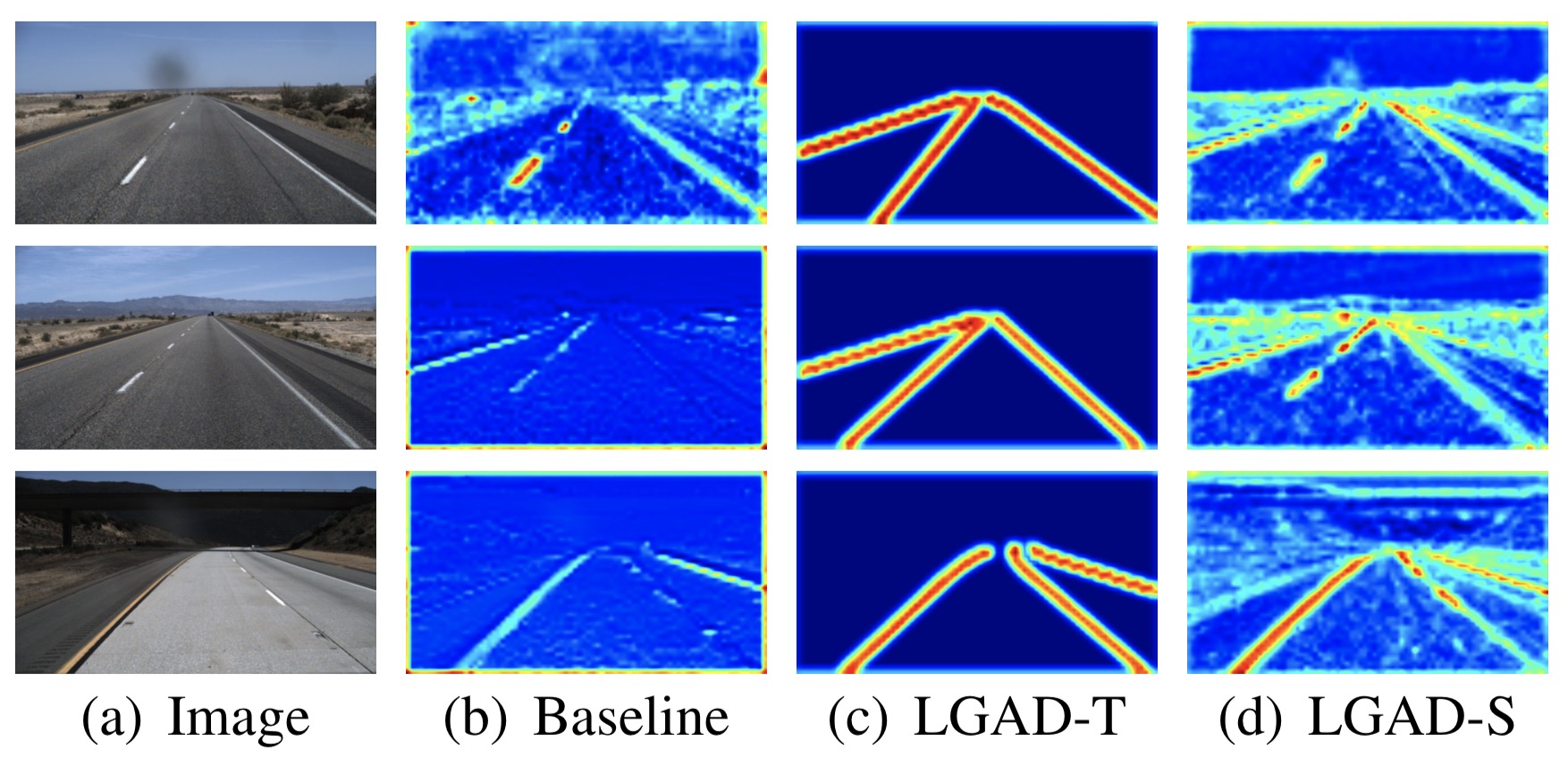}}
	\caption{Attention maps of PSPNet with and without LGAD. LGAD is adopted by the second stage output of the backbone to mimic the attention map of the teacher. (a) Input images. (b) Attention maps of the original PSPNet. (c) Attention maps of the LGAD teacher. (d) Attention maps of the LGAD supervised student. (best viewed in color)}
	\label{fig_bisnet}
\end{figure}

Several attempts have been made to address the aforementioned issues in lane segmentation. For instance, Pan \etal \cite{pan2018spatial} propose a SCNN scheme, in which slice-by-slice convolutions are used to enable message passing between neurons across rows and columns and better learn the continuous structure prior of lane markers. Neven \etal \cite{neven2018towards} use an additional H-Net to learn perspective transformation, conditioned on the input image. Then, the outputs of the lane instance segmentation network are transformed by the transformation matrix to get the fitting results. All these methods improve accuracy by adding complexity to neural network in both training phase and inference phase.

In this paper we aim to strengthen the long structure context capture ability of deep neural networks without sacrificing inference speed. We resort to utilizing lane labels more efficiently. We argue that the lane annotations can not only be used to supervise the loss of network outputs but also provide structure hints for the intermediate convolutional layers. Thus, we design a new distillation procedure which exploits label structure when training segmentation network. Specifically, for the purpose of broadcasting structure information throughout the network architecture, we convert the ground-truth label maps as images and then train a teacher from these images to replicate the images themselves. Then, the attention maps of the teacher are adopted to supervise the student lane segmentation network. The intuition is that the teacher network which receives input from a label to predict the label itself knows distinctly where its convolution layers should pay visual attention into. The student network benefits by mimicking the refined attention maps of the teacher network. We name this distillation mechanism as  \textit{Label-guided Attention Distillation} (LGAD). 

In contrast to conventional distillation, which starts with a powerful large teacher and performs knowledge transfer to a small student, LGAD starts with a teacher whose network structure remains the same with the student. This structure construction ensures accurate information transmission. It turns out that the student in such a distillation mechanism learns significantly better than when learning alone in a conventional supervised learning scenario. A comparison between the attention maps of PSPNet~\cite{zhao2017pyramid} before and after LGAD is illustrated in Figure \ref{fig_bisnet}. We let the second stage output of the backbone in PSPNet to mimic the attention map of the teacher. The LGAD mechanism is able to help the student network better capture long continuous contexts. As illustrated in the Figure \ref{fig_bisnet}(d), the student network shows much distinct attention at the lane locations under the LGAD mechanism. Note that all the four stages in PSPNet are benefited by the LGAD mechanism with richer scene contexts captured by the visual attention maps.

Moreover, the LGAD strategy is generally applicable, \eg, it can be easily incorporated in any lane segmentation network. As the teacher is deprecated after training, our method do not increase inference time, the performance improvements are essentially free. Note that this strategy offers the opportunity to train a small network with excellent performance. In our experiments, we demonstrate the effectiveness of LGAD on lightweight models, such as ENet \cite{paszke2016enet}. It turns out that lightweight ENet-LGAD surpasses existing lane segmentation methods.

Overall, different from the existing computing expensive techniques such as slide-by-slide convolution and perspective transformation learning, LGAD is a simple but efficient approach to train accurate land segmentation networks.  Our experiments show improvement from a variety of network architectures with LGAD embedded (\eg, BiseNet \cite{yu2018bisenet}, ENet \cite{paszke2016enet}, and PSPNet \cite{zhao2017pyramid}) and get consistent improvements on land segmentation. 

In this paper, our main contributions are as follows: 
\begin{itemize}
\item[(1)] We propose a novel distillation approach, \ie, LGAD, which uses structure hints in lane labels to guide the attention of lane segmentation networks to better capture long-range textural information. LGAD is only applied in the training phase and brings no computational cost in the inference phase.

\item[(2)] We carefully investigate the inner mechanism of LGAD, the considerations of choosing among different layer mimicking paths, and the optimization strategies.
\item[(3)] Extensive experiments on two benchmark datasets demonstrate the effectiveness of the proposed LGAD on boosting the performance of lane segmentation networks.

\end{itemize}

%-------------------------------------------------------------------------
\section{Related Work}

\textbf{Semantic segmentation.} Recent works in semantic segmentation always leverage fully-convolutional networks (FCNs) \cite{long2015fully}. U-Net \cite{ronneberger2015u} introduces skip connections between the feature maps of encoder and decoder. SegNet \cite{badrinarayanan2017segnet} stores pooling indices and reuses them in the decoder part. RefineNet \cite{lin2017refinenet} generics multi-path refinement network that using long-range residual connections to enable high-resolution prediction. A limitation of these FCN based networks is their small receptive field which prevents them from taking large scope contexture information into account. The Deeplab series of methods \cite{chen2014semantic,chen2017rethinking,chen2017deeplab} resort to dilated convolution and atrous spatial pyramid pooling to increase the receptive field.  Deeplab-v3+~\cite{chen2018encoder} attaches the decoder part after the atrous spatial pyramid pooling module to extracts rich semantic features. PSPNet \cite{zhao2017pyramid} improves the performance via pyramid pooling module, which extracts global context information by aggregating different region-based contexts. BiSeNet~\cite{yu2018bisenet} combines a high-resolution spatial path and a fast downsampling context path to make the right balance between efficiency and segmentation performance. These networks have the ability to model relatively large range of contextual relationships. However, the resulting feature representation is dominated by large objects, and consequently, performance on small-size objects is limited.

Mostajabi \etal \cite{mostajabi2018regularizing} propose to enhance label structure information for semantic segmentation. They first train an autoencoder on the ground-truth labels map aim at compressing the label information into the bottleneck abstract representation. Then a typical CNN backbone (\eg VGG-16 or DenseNet) is connected to the pre-trained and frozen decoder to form the segmentation network. The applicability of this strategy is limited to the encoder-decoder segmentation structure. However, LGAD is generally applicable and can be used to boost the performance of any type of lane segmentation network. Moreover, LGAD focus on transferring attention maps which are more flexible than abstracted representation.  

\begin{figure*}[t]
	\centering
	\includegraphics[width=14.3cm]{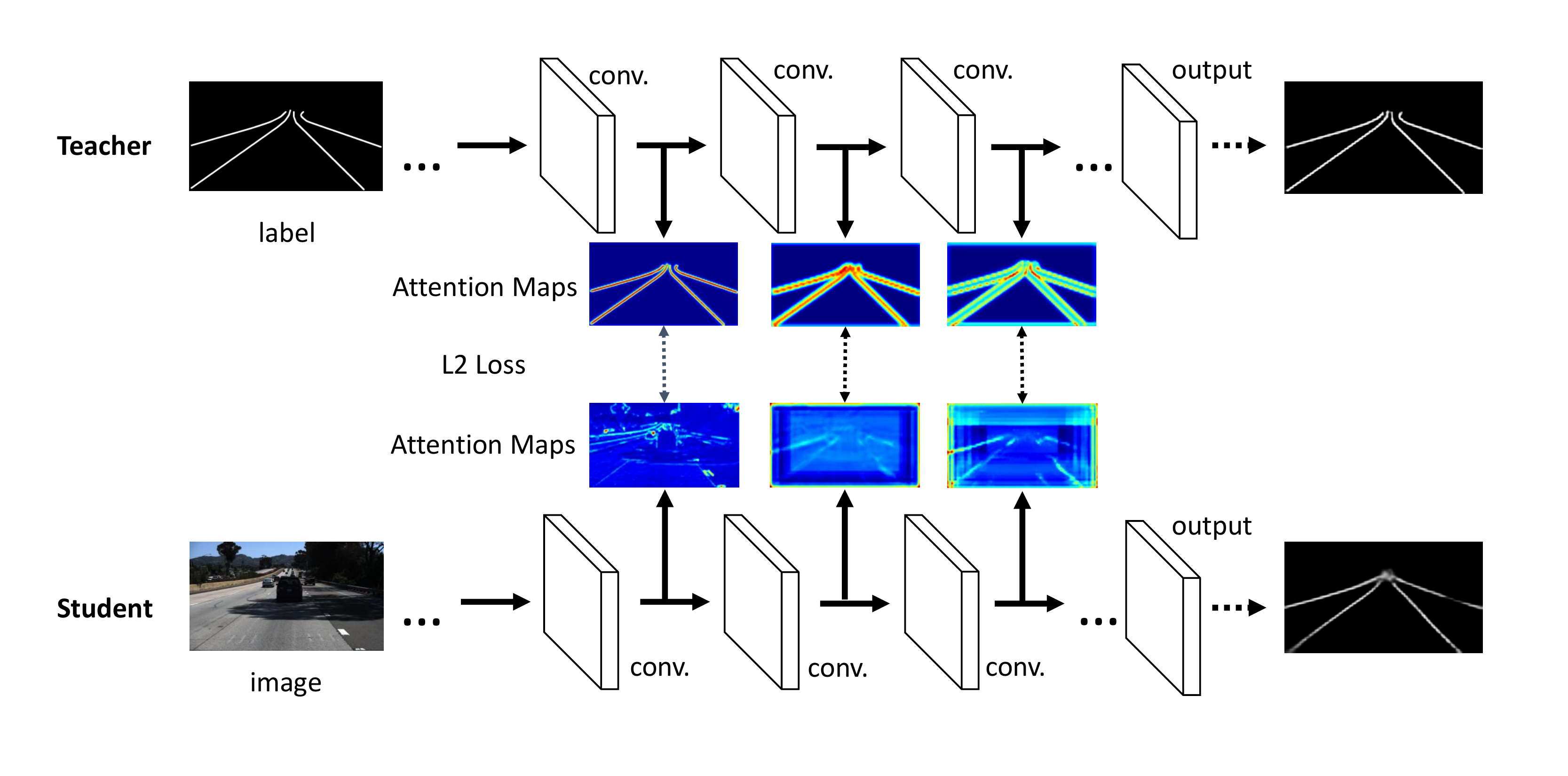}
	\caption{Schematics of the proposed LGAD method. The teacher network is trained from input labels to predict the labels themselves. Then, the attention maps of the teacher are transferred to student, making the student network not only make correct predictions but also mimic the desirable attention. The two networks share the same structure. The inputs of the teacher network and student network should be paired.}
	\label{fig_roadmap}
\end{figure*}

\textbf{Lane Segmentation.} Traditional lane segmentation methods are mainly based on hand-crafted  features~\cite{son2015real,jung2015efficient}. Hough transform~\cite{liu2010combining} and Kalman filters~\cite{zhou2010novel} are popular choices to assemble the low-level features to form the lanes. These methods always suffer from lacking robustness.

In recent years, several attempts have been made to apply deep learning to lane detection~\cite{pan2018spatial,ghafoorian2018gan,neven2018towards}. 
SCNN~\cite{pan2018spatial} generalizes traditional deep layer-by-layer convolutions to slice-by-slice convolutions within feature maps, thus enabling message passings between pixels across rows and columns in a layer. Neven \etal \cite{neven2018towards} use a regular segmentation network to get lane marking prediction maps with a second network to perform a perspective transformation, after which curve fitting is used to obtain the final results. Ghafoorian \etal \cite{ghafoorian2018gan} proposes
embedding-loss GAN (EL-GAN) for lane segmentation, where the lanes were predicted by a generator based on the input image, and judged by a discriminator with shared weights. Li \etal \cite{li2017deep} divide a road image into a number of continuous slices. Then, a recurrent neural network is used to infer the lanes from CNN extracted features on the image slices. While the aforementioned approaches do improve lane segmentation performance, the additional mechanisms (\eg layer-by-layer convolution, learned perspective transformation, GAN based label-resembling, and RNN procedure) are short of efficiency. 

Different from the above deep learning-based methods, the proposed LGAD resorts to a distillation mechanism to reinforces lane information in the convolutional feature maps. Meanwhile, LGAD can be seamlessly integrated into any CNN based segmentation framework to improve performance, and it not increase the inference time. 

\textbf{Knowledge Distillation.} Knowledge Distillation (KD) is proposed to compress huge models into simpler and faster models. In this context, Ba and Caruana~\cite{ba2014deep} first introduce knowledge distillation by minimizing L2 distance between the features from the last layers of two networks. Later, Hinton \etal \cite{hinton2015distilling} show that the predicted class probabilities from the teacher are informative for the student and can be used as a supervision signal in addition to the regular labeled training data during training. Czarnecki \etal \cite{czarnecki2017sobolev} minimize the difference between the teacher and the student derivatives of the loss combined with the divergence from the teacher predictions. Heo \etal \cite{heo2019knowledge} explore knowledge distillation using activation maps. Zagoruyko \etal \cite{zagoruyko2016paying} propose to apply knowledge distillation to internal attention maps. They improve the performance of a student CNN network by forcing it to mimic the attention maps of a powerful teacher network.  The proposed LGAD differs from \cite{zagoruyko2016paying} in that the goal of our distillation mechanism is to transfer structure label information rather than the representation ability of a powerful teacher. Specifically, the LGAD teacher shares the same structure with the student and is trained in a label-to-label manner. It is noteworthy that our focus is to investigate the possibility of distilling label guided attention.

Hou \etal \cite{hou2019learning} follow the same idea to use a distillation strategy to train lightweight lane segmentation models. They propose a Self Attention Distillation (SAD) approach, which uses attention maps from higher layers as the distillation targets for its lower layers. Compared with SAD, we argue that the distillation target of LGAD is more reasonable and accurate. Conventional, the lower layers and higher layers in a CNN play different roles, \ie, capture local features and capture rich contextual features. Different layers should inherently have different vision attentions. In LGAD, each teacher layer knows exactly where the feature maps should pay attention to at its post. With superior attention information, ENet-LGAD surpasses ENet-SAD on the benchmarks (97.65\% vs 96.64\% on TuSimple, 72.0\% vs 70.8\% on CULane).

\begin{figure}[b]
	\centering
	\subfigure[input]{\begin{minipage}[t]{0.282\linewidth}
			\includegraphics[width=0.9in]{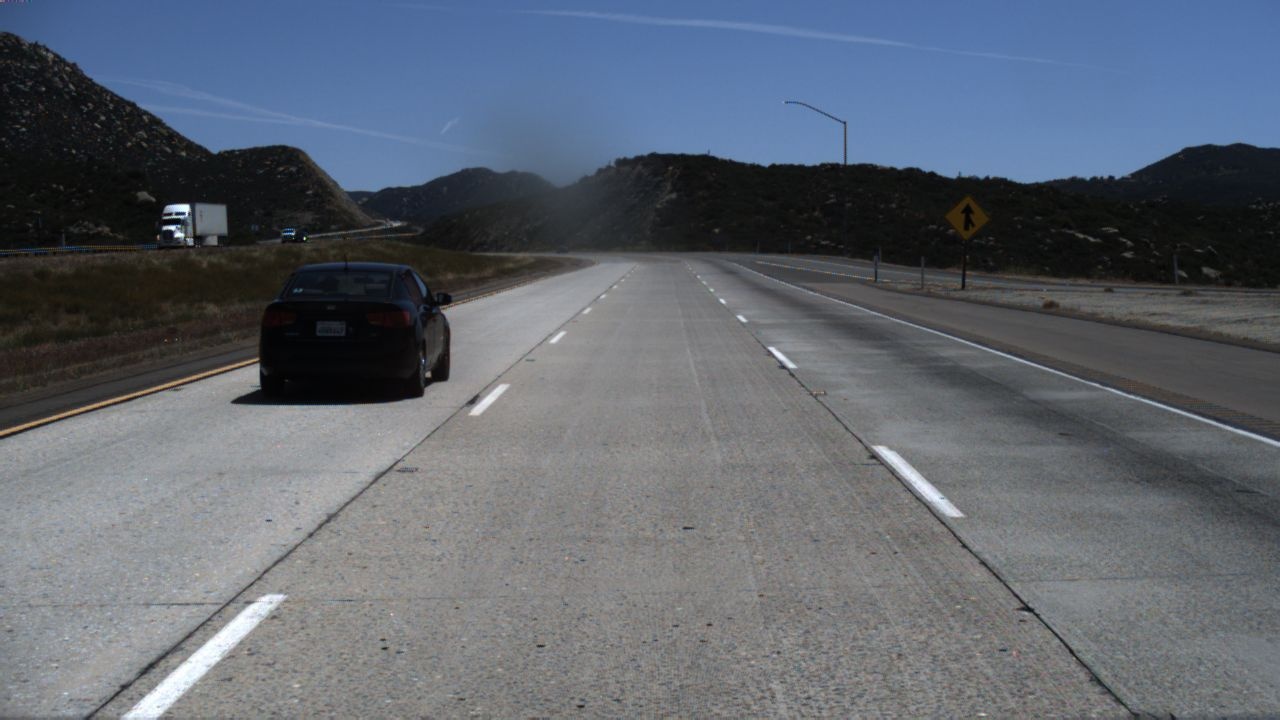}\vspace{2pt}
	\end{minipage}}
	\subfigure[${F_{mean}}(R_m)$]{\begin{minipage}[t]{0.282\linewidth}
			\includegraphics[width=0.9in]{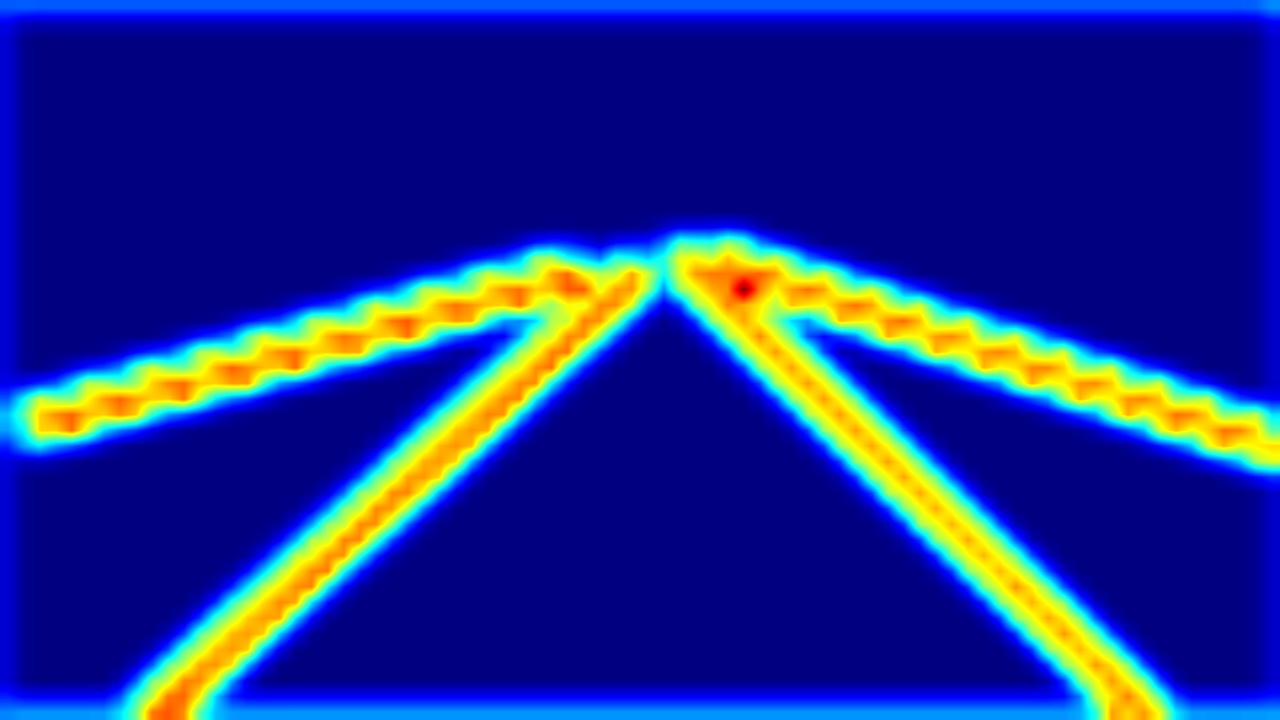}\vspace{2pt}
	\end{minipage}}
	\subfigure[${F^{2}_{mean}}(R_m)$]{\begin{minipage}[t]{0.282\linewidth}
			\includegraphics[width=0.9in]{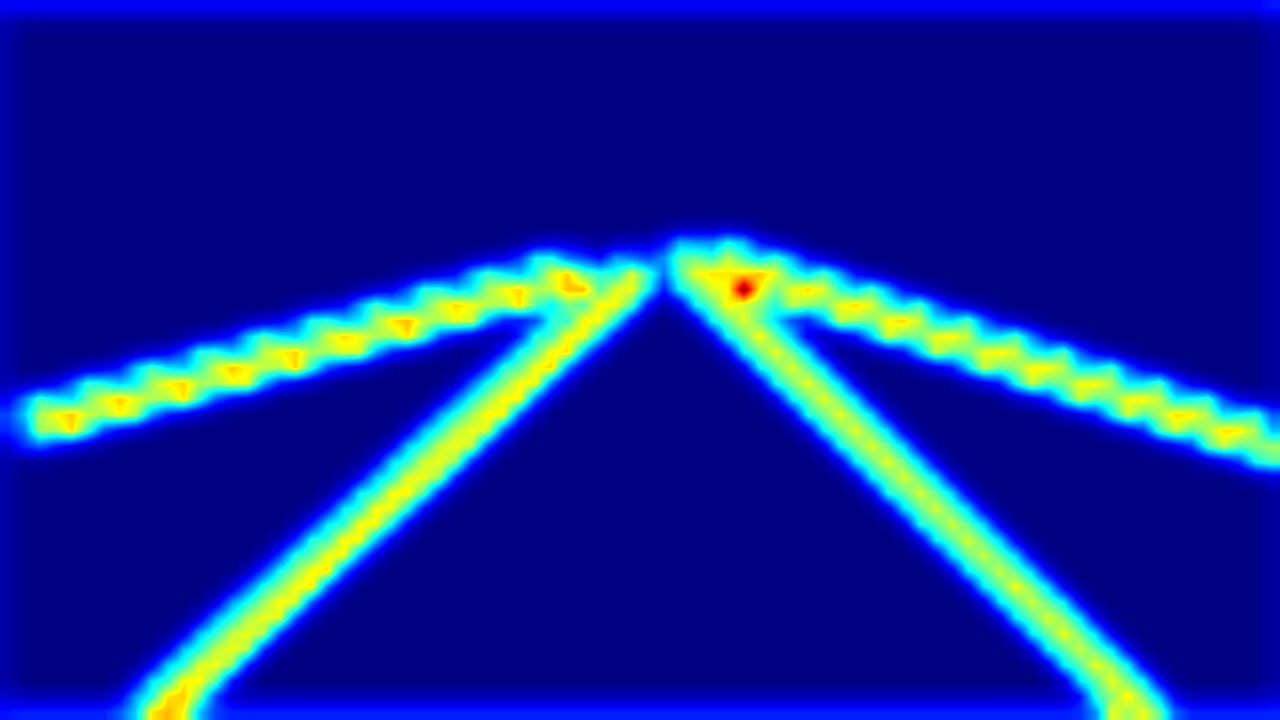}\vspace{2pt}
	\end{minipage}}
	%     \subfigure[${F_{max}}(R_m)$]{\begin{minipage}[t]{0.242\linewidth}
	%     			\includegraphics[width=0.7in]{mean_max/max/101_test_label.jpg}\vspace{2pt}
	%     	\end{minipage}}
	\caption{Attention maps of the second stage layer in the PSPNet using different map functions.}
	\label{fig_attention}
\end{figure}

\begin{figure*}[t]
	\centering
	\centerline{\includegraphics[width=12.3cm]{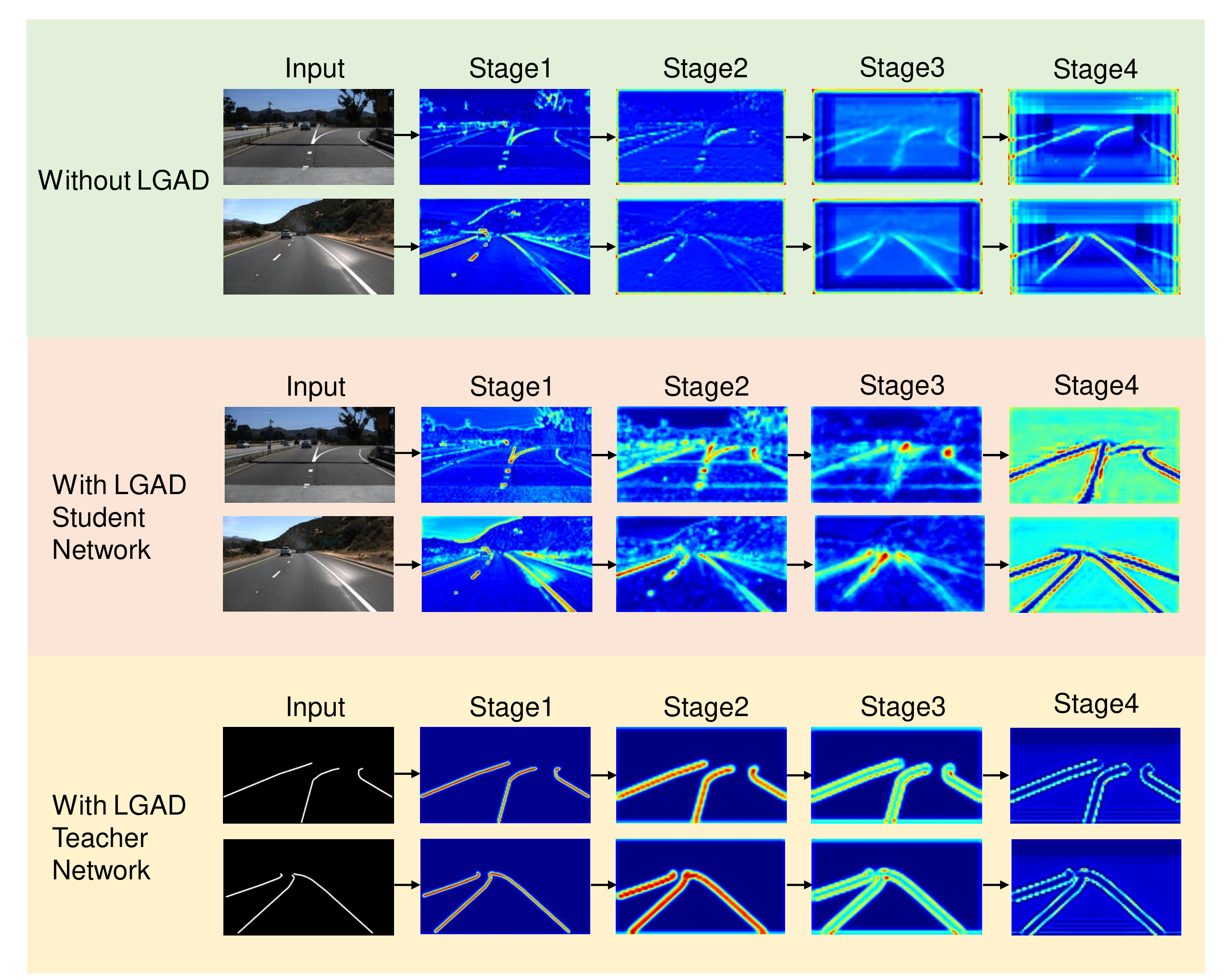}}
	\caption{Attention maps for four stages of the backbone in the PSPNet with and without LGAD. Both networks are trained in TuSimple dataset up to 50 epochs. (best viewed in color)}
	\label{fig_pspatt}
\end{figure*}

\section{Method}

Conventionally, a lane segmentation network is trained in an end-to-end manner, with the ground-truth labels to compute the loss function appended at the end of the network. We show that detailed ground-truth lane annotations contain additional information that existing schemes for training lane segmentation networks fail to exploit. By designing a knowledge distillation training procedure, we are able to capture lane structure information, and as a result boost performance at test phase. The proposed \textit{label-guided attention distillation} (LGAD) method contains two distinct network modules, \ie, a teacher network, and a student network. The teacher network models the structure of the labels themselves and performs layer-wise attention distillation to the student segmentation network. Figure \ref{fig_roadmap} illustrates this scheme. In the following, we explain how we perform layer-wise attention distillation. Then, we elaborate on how to add LGAD to network training as well as the optimization strategies.

\subsection{Activation-based Attention}

In general, attention maps can be categorized into two types: \textit{activation-based} attention map and \textit{gradient-based} attention map \cite{zagoruyko2016paying}. The activation-based attention map encodes what spatial areas of network focus for generating output, while the gradient-based attention map encodes how sensitive to the feature the output prediction is. As LGAD aims to transfer spatial structure hints in the labels as well as teach the student layers what areas the attention should pay into, activation-based attention is a more rational choice than gradient-based attention. Moreover, the activation-based attention transition yields consistently performance gains in distillation methods while gradient-based attention distillation barely works \cite{zagoruyko2016paying, hou2019learning}. So we only discuss the activation-base attention distillation as follows.

The activation-based attention map is computed by processing the absolute values of the elements of a tensor. We denote the tensor of the \textit{m}-th CNN layer as $R_m^{{C_m} \times {H_m} \times {W_m}}$, which consists of $C_m$ feature channels with spatial dimension $H_m \times W_m$. The attention map generation function $\mathcal{F}$ takes the above 3D activation tensor $R_m^{{C_m} \times {H_m} \times {W_m}}$ as input and generates a channel-flattened 2D attention map $A_m^{{H_m} \times {W_m}}$:
\begin{equation}\label{equation1}
\mathcal{F} : R_m^{{C_m} \times {H_m} \times {W_m}} \to A_m^{{H_m} \times {W_m}}.
\end{equation}
The absolute value in each neuron activation of $R^{{C_m} \times {H_m} \times {W_m}}$ indicates the importance of this neuron in the corresponding CNN layer. Therefore, the attention maps can be constructed by computing statistics of these values across the channel dimension:

\begin{itemize}
	\item mean of absolute values:$\mathcal{F}_{mean}(R_m) = \frac{1}{C}\sum\limits_{i = 1}^C {{{\left| {{R_{mi}}} \right|}}} $
	\item mean of absolute values raised to the power of $p$ (where $p > 1$): $\mathcal{F}_{mean}^p(R_m) = \frac{1}{C}\sum\limits_{i = 1}^C {{{\left| {{R_{mi}}} \right|}^p}} $
	%\item max of absolute values raised to the power of $p$ (where $p > 1$): $\mathcal{F}_{max}^p(R_m) = {\max _{i = 1,C}}{\left| {{R_{mi}}} \right|^p}$
\end{itemize}
where $R_{mi}$ denotes the $i$-th slice in $R_{m}$ in the channel dimension, and the mean, power and absolute value operations are element wise. 
We illustrate the attention maps generated by these functions on the last pre-downsampling feature layer of PSPnet in Figure \ref{fig_attention}. It can be seen that these functions have different properties. Compares to ${\mathcal{F}_{mean}}(R_m)$, $\mathcal{F}_{mean}^p(R_m)$ (where $p > 1$) puts more weights on the spactial locations with higher activations. 

In the experiment, $\mathcal{F}_{mean}(R_m)$ instead of $\mathcal{F}_{mean}^p(R_m)$ exhibits superior performance. We think one possible cause is that the boundaries of attention areas should be slightly expanded for dealing with occlusion and missing.

\subsection{Label-guided Attention Distillation}

LGAD transfers attention from a teacher network to a student network. Different from original knowledge distillation methods which always has a powerful teacher and small student, the teacher and the student share the same structure in LGAD strategy. The teacher network captures label structure information by replicating the input label maps as outputs. The intuition behind LGAD is that the label-to-label trained teacher knows where to pay attention. The student network will benefit from not only making correct predictions but also having attention maps that are similar to the teacher.

LGAD can be easily added to an exiting segmentation network. Figure \ref{fig_roadmap} is a schematic of LGAD method. The teacher network takes label image as inputs and tries to predict the label itself. The transferring losses are placed between teacher and student attention maps generated from the same stage to ensure accurate information transmission. 
Let $S$, $T$, $R^S$, $R^T$ denote student, teacher, and their weights correspondingly. Let $\mathcal{N}$ denote the indices of all the teacher-student activation layer pairs between which we want to transfer attention maps. Then the layer-wise distillation loss ${\mathcal{L}_{AT}}$ is computed as follows: 
\begin{equation}\label{equation2}
{\mathcal{L}_{AT}}(S,T) = \sum\limits_{j \in \mathcal{N}} {{\mathcal{L}_d}(F_{mean}(R_j^S) - F_{mean}(R_j^T))} 
\end{equation}
where ${\mathcal{L}_d}$ is typically defined as ${\mathcal{L}_2}$ loss. In the example illustrated in Fig.~\ref{fig_roadmap}, the number of layer pairs $\mathcal{N} = 3$. As a result, the final objective function is as follow
\begin{equation}\label{equation3}
\mathcal{L} = {\mathcal{L}_{seg}}(s,\hat s) + \alpha {\mathcal{L}_{AT}}(S,T)
\end{equation}
where ${\mathcal{L}_{seg}}$ is the standard cross entropy loss. $\hat s$ is the segmentation map generated by the network. The 
$\alpha$ is loss coefficient which is set to $0.5$ in the experiments.

Figure \ref{fig_pspatt} visualized the attention maps for three layers in PSPNet with and without LGAD. As can be seen that, when trained with LGAD, the attention maps of the PSPNet concentrates more on lane locations. This would significantly improve the segmentation performance, as we will report in the experiments.

\subsection{Optimization}

The proposed LGAD can be optimized by two different strategies. One is that the teacher and the student are optimised jointly and collaboratively. In this way, the LGAD strategy is embedded into each mini-batch based model update step for both networks and throughout the whole training process. At each iteration, the two networks are learning with the same batch, the teacher is trained with standard segmentation loss ${\mathcal{L}_{seg}}(s,\hat s)$ and the student is trained with the loss function defined in Equation~\ref{equation3}. The other strategy is that we follow the original distillation procedure, and make the one-way attention transfer between a static pre-trained teacher and a fresh student. We compare the performances of these two strategies in Section \ref{exp_ablation}. We empirically find that these two strategies have almost equivalent performance. So, we only provide results based on the original distillation optimization strategy in 
the other experiments for convenience. The optimization procedure is detailed in Algorithm \ref{alg:Framwork}. 

\begin{algorithm}[h] 
	\caption{ Label-guided Attention Distillation.} 
	\label{alg:Framwork} 
	\begin{algorithmic}[1]
		\REQUIRE The lane dataset $\mathcal{D} = \{\textbf{x}, \textbf{y}\}$, where $\textbf{x}$ denote scene images and $\textbf{y}$ denote the corresponding labels; The loss coefficient $\alpha $;
		\STATE Initialize the teacher network $\mathcal{N}_{te}$;
		\REPEAT
		\STATE Random select a batch of labels $\{\textbf{y}_i\}_{i=1}^m$;
		\STATE Employ the teacher network $\textbf{y}_{i}\leftarrow \mathcal{N}_{te}\left (\textbf{y}_i \right )$; \\
		\STATE Calculate the cross entropy loss $\mathcal{L}_{seg}$;
		\STATE Update weights in $\mathcal{N}_{te}$ according to the gradient;\\
		\UNTIL{convergence} \\
		\STATE Initialize the student network $\mathcal{N}_{st}$, whose structure keep the same with $\mathcal{N}_{te}$;
		\REPEAT
		\STATE Random select a batch $\{\textbf{x}_i,\textbf{y}_i\}_{i=1}^m$;
		\STATE Employ the teacher and student network:\\ $\textbf{y}_{i}\leftarrow \mathcal{N}_{te}\left (\textbf{y}_i \right )$; $\textbf{y}_{i}\leftarrow \mathcal{N}_{st}\left (\textbf{x}_i \right )$; \\
		\STATE Calculate the attention distillation loss in Eq.~\ref{equation2};
		\STATE Calculate the total loss in Eq.~\ref{equation3};
		\STATE Update weights in $\mathcal{N}_{st}$ using back-propagation;\\
		\UNTIL{convergence} \\
		\ENSURE The student network $\mathcal{N}_{st}$.\\
	\end{algorithmic}
\end{algorithm}

It is possible to place LGAD between any correlated layer-pairs of the teacher and the student, which could affect the performance. In the LGAD process, distillation can be applied to multi-layers or a single layer. In general, the number of possible LGAD position for a network with $n$ layers is $(C_n^1 + C_n^2 +  \cdots C_n^n) = 2^n-1$. We will discuss the impact of distillation position in Section \ref{exp_ablation}.

\begin{figure}[t]
	\centering
	\subfigure{\begin{minipage}[t]{0.31\linewidth}
			\includegraphics[width=1in]{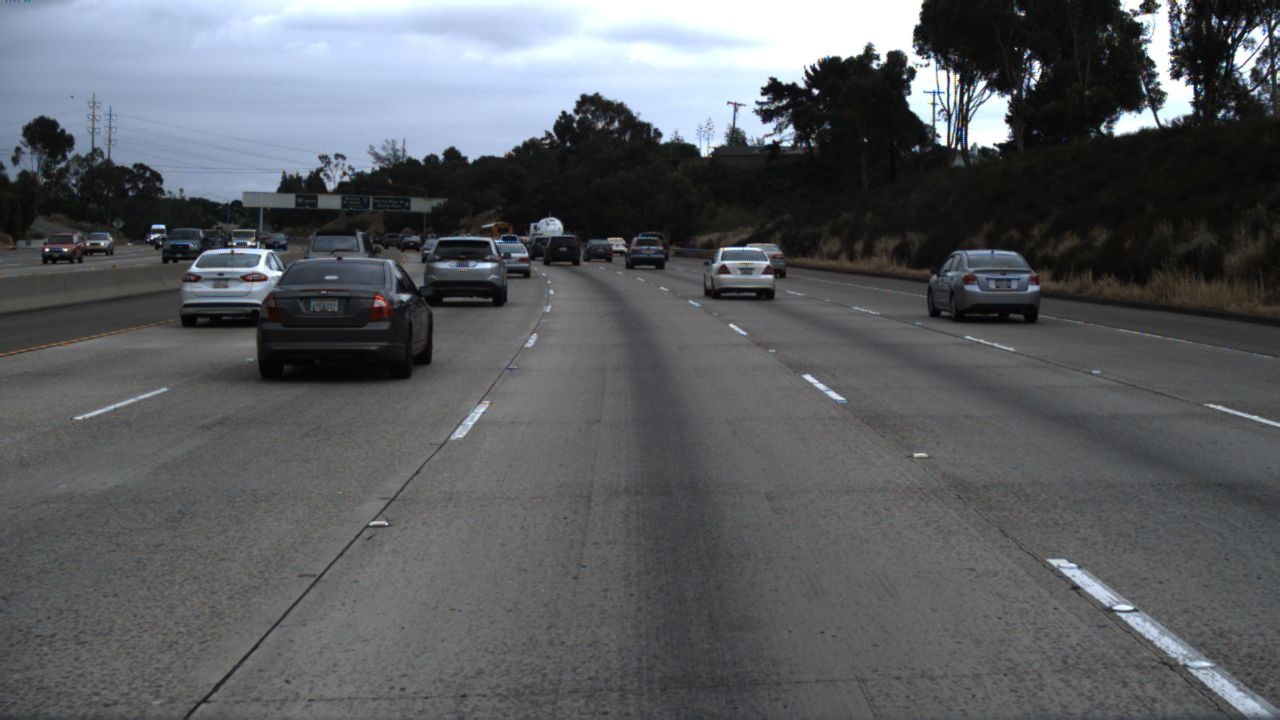}\vspace{5.7pt}
			\includegraphics[width=1in]{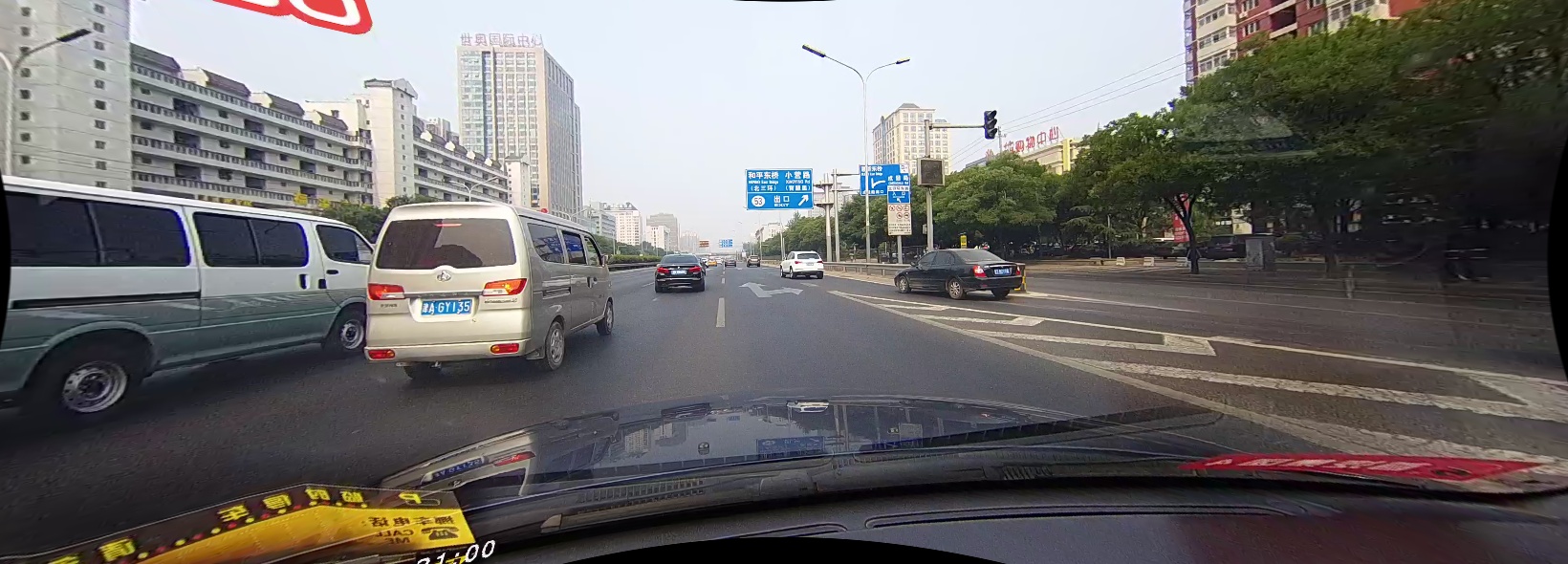}\vspace{6pt}
	\end{minipage}}
	\subfigure{\begin{minipage}[t]{0.31\linewidth}
			\includegraphics[width=1in]{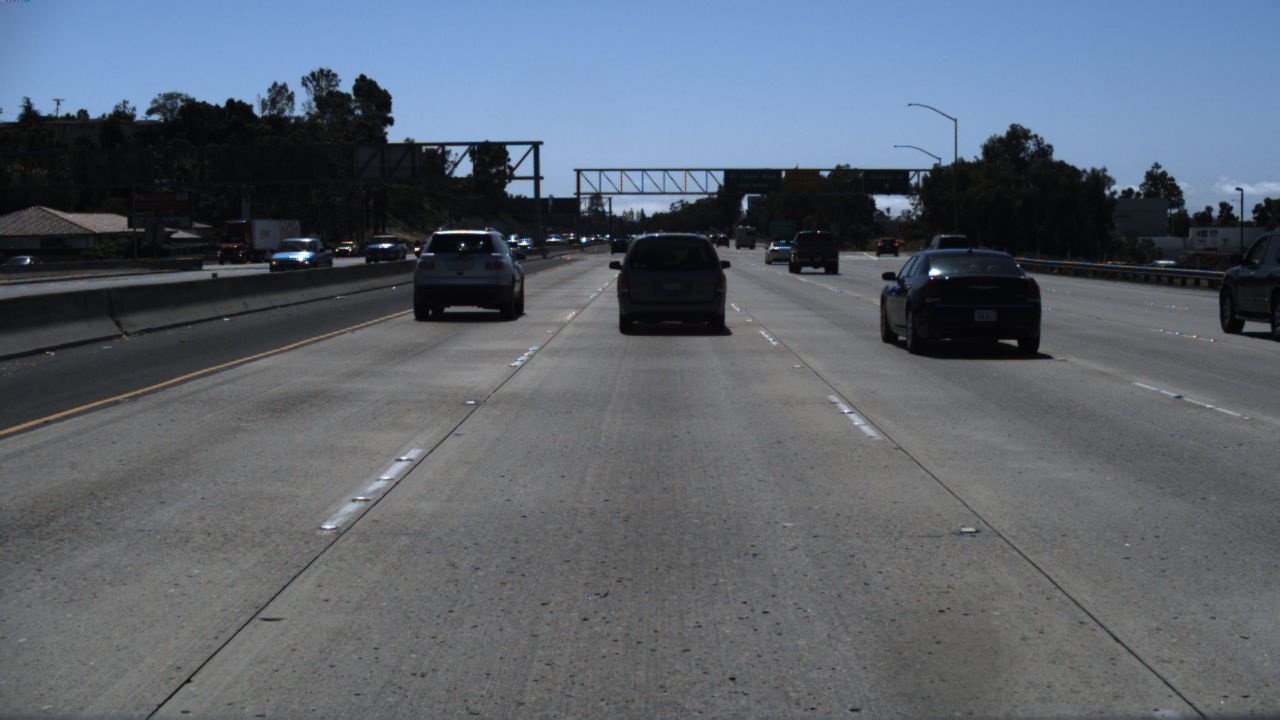}\vspace{5.7pt}
			\includegraphics[width=1in]{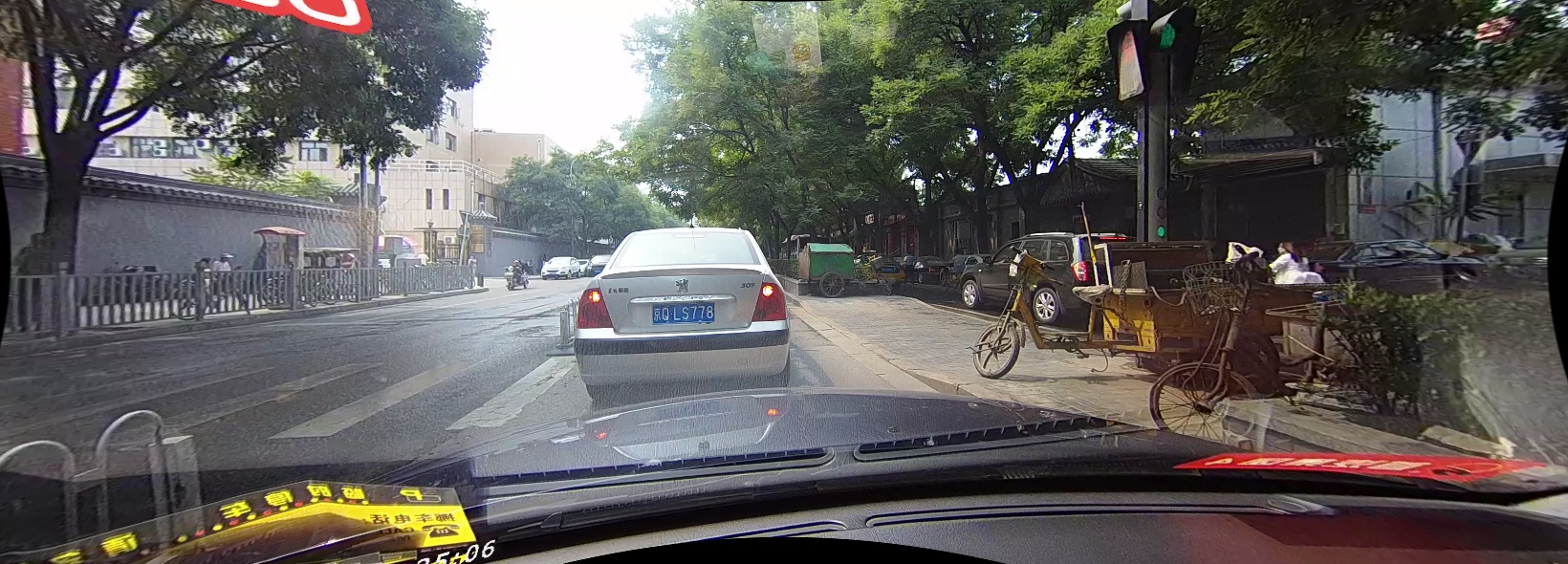}\vspace{6pt}
	\end{minipage}}
	\subfigure{\begin{minipage}[t]{0.31\linewidth}
			\includegraphics[width=1in]{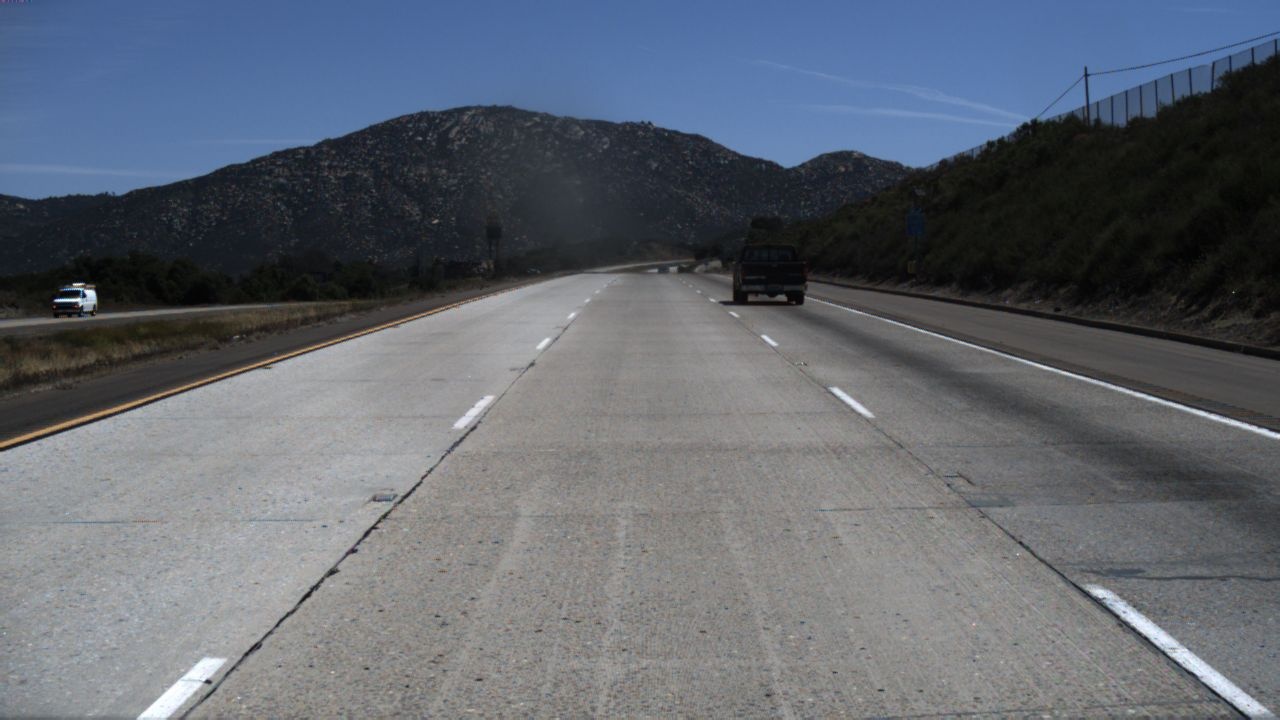}\vspace{5.7pt}
			\includegraphics[width=1in]{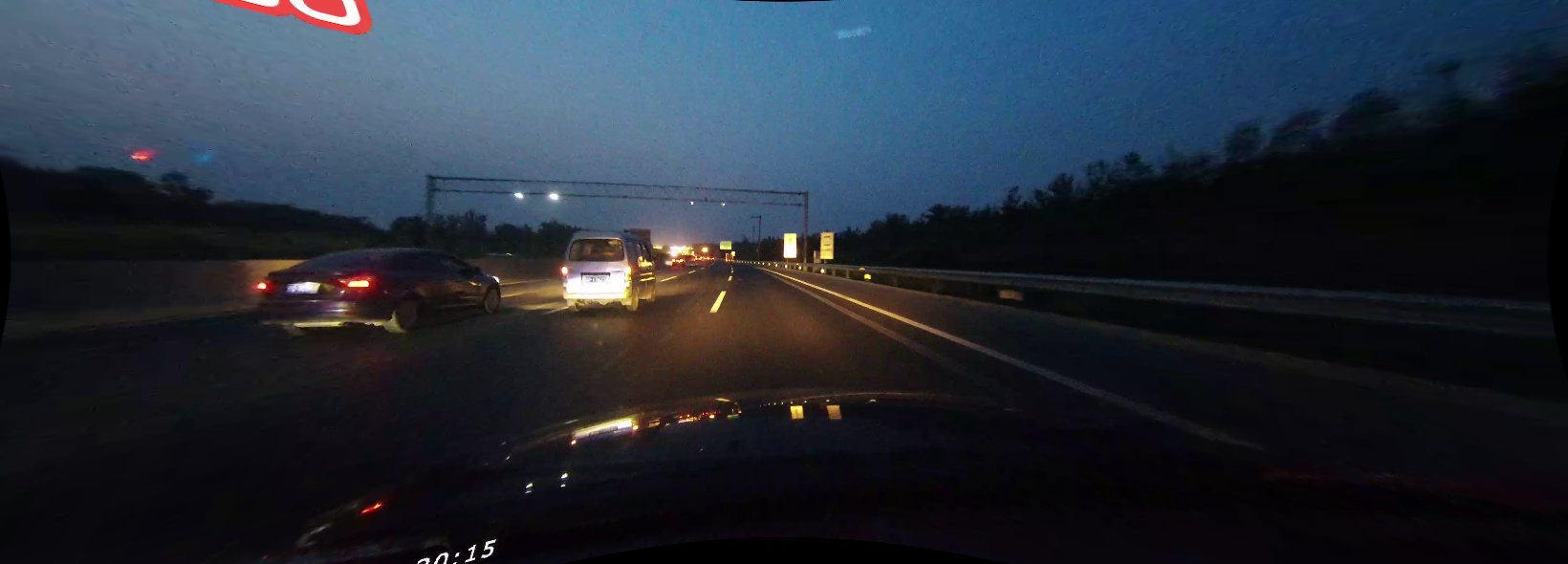}\vspace{6pt}
	\end{minipage}}
	\caption{Sample images of lane segmentation datasets. Top row: images from TuSimple. Bottom row: images from CULane.}
	\label{fig_datasets}
\end{figure}

\section{Experiments}

We first give an overview of the datasets, evaluation metrics, and the implementation details used in the experiments. Then we present the main results and ablation study. 

\subsection{Datasets and Setup}
We validate LGAD on two popular benchmarks, \textit{i.e.}, TuSimple~\cite{TuSimple} and CULane~\cite{pan2018spatial}. Sample frames are illustrated in Figure \ref{fig_datasets}.

\textbf{TuSimple dataset.} This is the largest lane segmentation dataset before 2018. This dataset consists of 6,408 labeled images, which are split into 3,626 training images and 2,782 test images. These images are captured under good and medium weather conditions as well as different traffic environments with resolution $1280 \times 720$. They contain highway roads with a different number of lanes, such as 2 lanes, 4 lanes, or more. For each image, 19 previous frames are also provided, but without annotation. We only use the annotated images in our experiments.

\textbf{CULane dataset.} The CULane dataset is comprised of 133,235 frames extracted from 55 hours of video. They divide the dataset into 88,880 training images, 9,675 validation images, and 34,680 test images. These images are undistorted and have a resolution of $1640 \times 590$. The test set is further split into normal and other challenging categories, including crowded and shadow scenes. It is a challenging dataset considering the total number of video frames and the road types. 

\textbf{Lane prediction process.} For the CULane dataset, lane detection requires precise prediction of curves. We follow the original scheme in~\cite{pan2018spatial} to predict the existence of land markings,~\ie, the probability maps are sent to a small network, which contains AvgPooling and FC layers, to get the existence vector. Then, for each lane mark whose existence value is larger than $0.5$, we search the corresponding probability map every 20 rows for the position with the highest response value. At last, cubic splines are used to connect these positions to get the output curves. For the TuSimple dataset, the output probability maps of the segmentation models are not post-processed.

\textbf{Evaluation metrics.} 
In order to compare the performance to other competitors in TuSimple and CULane, we calculate the accuracy using the official evaluation metrics of TuSimple and CULane proposed in the literature.

(1) \textit{TuSimple.} We calculate the accuracy using Tusimple metrics. The prediction accuracy is computed as:
\begin{equation}\label{accuracy} 
accuracy = \frac{{{N_{correct}}}}{{{N_{gt}}}}
\end{equation}
where $N_{correct}$ is the number of correctly predicted points and $N_{gt}$ is the number of ground truth points. A point is considered correct if the predicted point and a ground truth point is within a certain threshold. Besides, we also report the false positive (FP) rate and the false negative (FN) rate.

(2) \textit{CULane.} We follow the standard process in~\cite{pan2018spatial} to judge whether a lane marking is correctly detected. We treat lane markings as lines with 30-pixel width and calculate the intersection-over-union (IoU) between the ground truth points and the predicted points. Predictions whose IoUs are larger than $0.5$ is considered as true positive (TP). The F-measure is employed as the final evaluation metric, which is computed as:
\begin{equation}
{F_{_\beta }} = (1 + {\beta ^2})\frac{{Precision \times Recall}}{{{\beta ^2}Precision + Recall}}
\end{equation}
where $Precision = \frac{{TP}}{{TP + FP}}$ and $Recall=\frac{TP}{TP+FN}$. Here $\beta$ is setting to 1, corresponding to harmonic mean (F1-measure). 

\subsection{Implementation Details.}
We perform experiments on ENet \cite{paszke2016enet}, PSPNet \cite{zhao2017pyramid} and BiseNet \cite{yu2018bisenet} to evaluate the efficiency of the proposed method as they are three typical segmentation models of different structures. ENet is a classical Encoder-Decoder model, PSPNet is a model without a decoder and BiseNet utilizes a compound structure. The weights of both the teacher and the student are initialized by training the networks on the ImageNet \cite{ILSVRC15}.
%We train the teacher and student networks from scratch in the experiments.
We adopt the Stochastic Gradient Descent (SGD) update rule \cite{bottou2010large} with the aforementioned loss function in Equation \ref{equation3} to optimize the parameters of the network. Loss coefficient $\alpha$ is set to $0.5$.

We adopt the ‘poly’ policy widely used in a lot of segmentation models to set the learning rate for each iteration, where the learning rate in an iteration equals to initial learning rate multiplied by $(1-\frac{iter}{max_iter})^{power}$. The power is set to 0.9 and the initial learning rate is set to 0.025 for student network and 0.007 for teacher network. All of the teacher networks are trained by 50 epochs. The student networks are trained by 150 epochs for TuSimple and 100 epochs for CULane. 

We also apply data augmentation to avoid overfitting. Similar to~\cite{zhao2017pyramid}, random cropping, horizontal flipping, and random rotation are adopted to preprocess the input images. We apply the same data argumentation and segmentation losses to BiseNet baselines, PSPNet baselines, ENet baselines, and deep distillation processes for a fair comparison. Most Experiments are done on four GTX 2080Ti GPUs with batch size of 8 except the experiment to record runtime, which is done on a single GTX TITAN X GPU for a fair comparison.

\begin{figure}
	\centering
	\subfigure[Input]{\begin{minipage}[t]{0.22\linewidth}
			\includegraphics[width=0.7in]{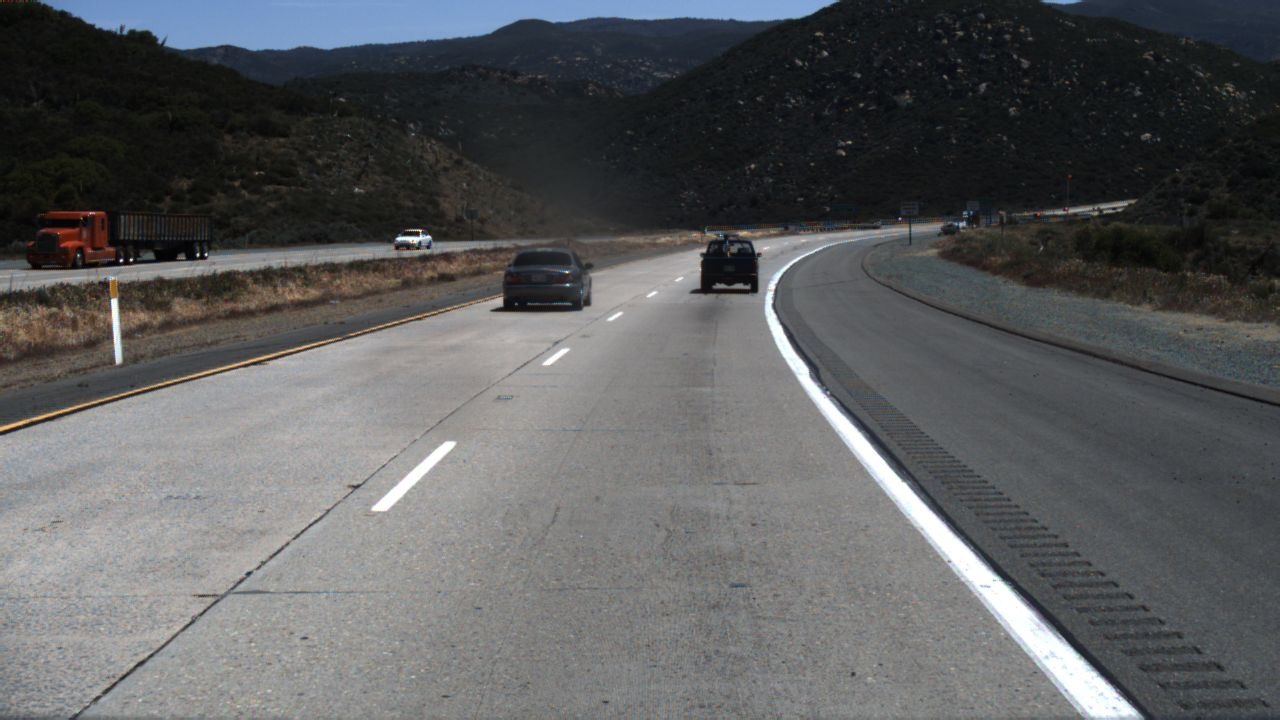}\vspace{3pt}
			\includegraphics[width=0.7in]{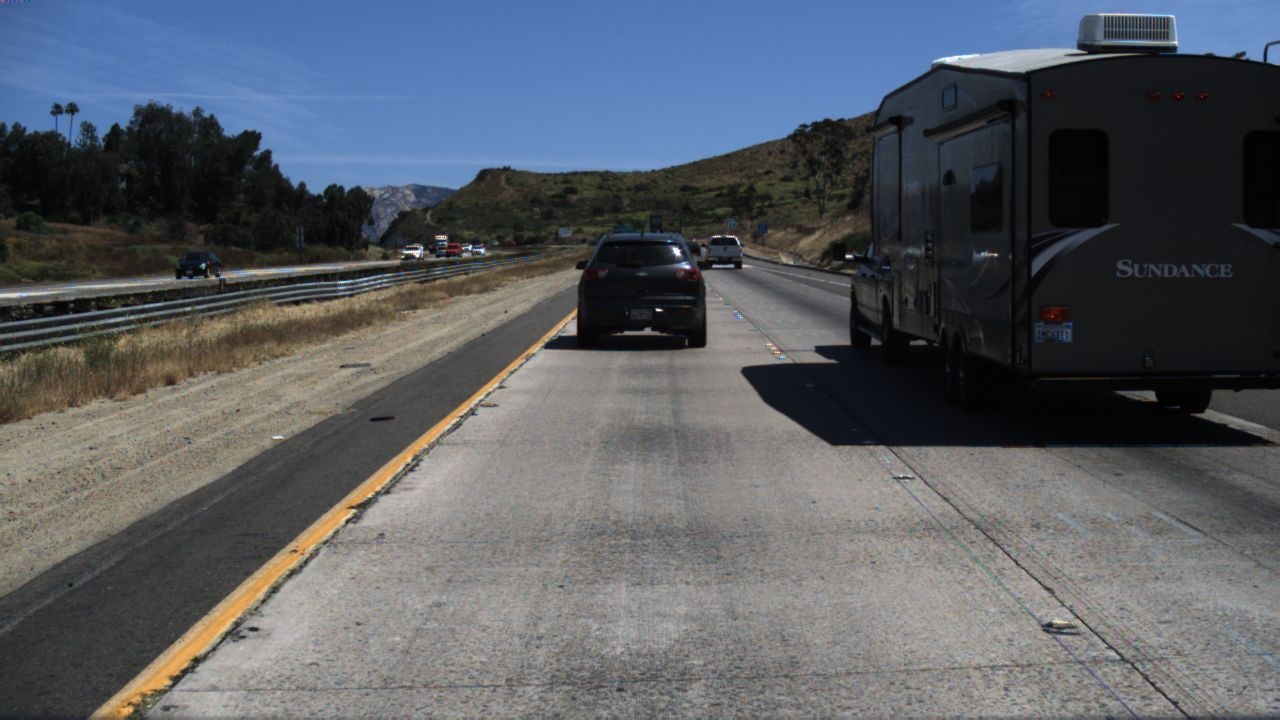}\vspace{3pt}
			\includegraphics[width=0.7in]{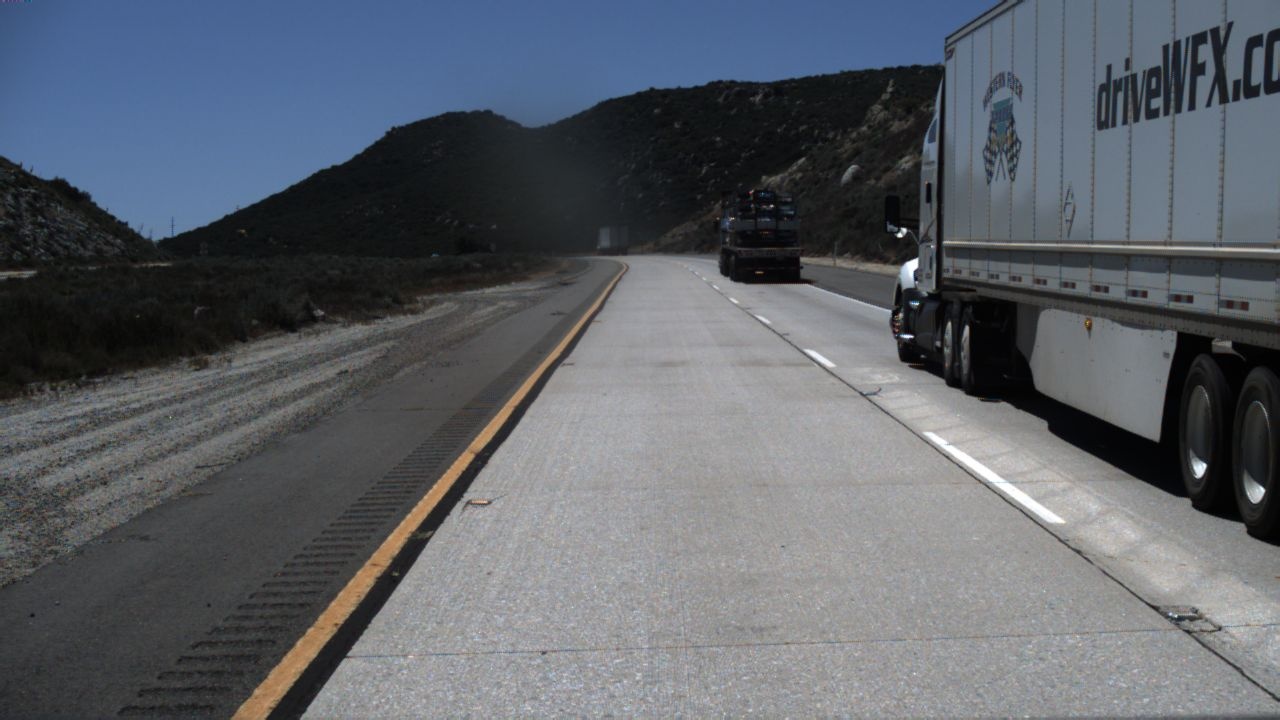}\vspace{3pt}
			\includegraphics[width=0.7in]{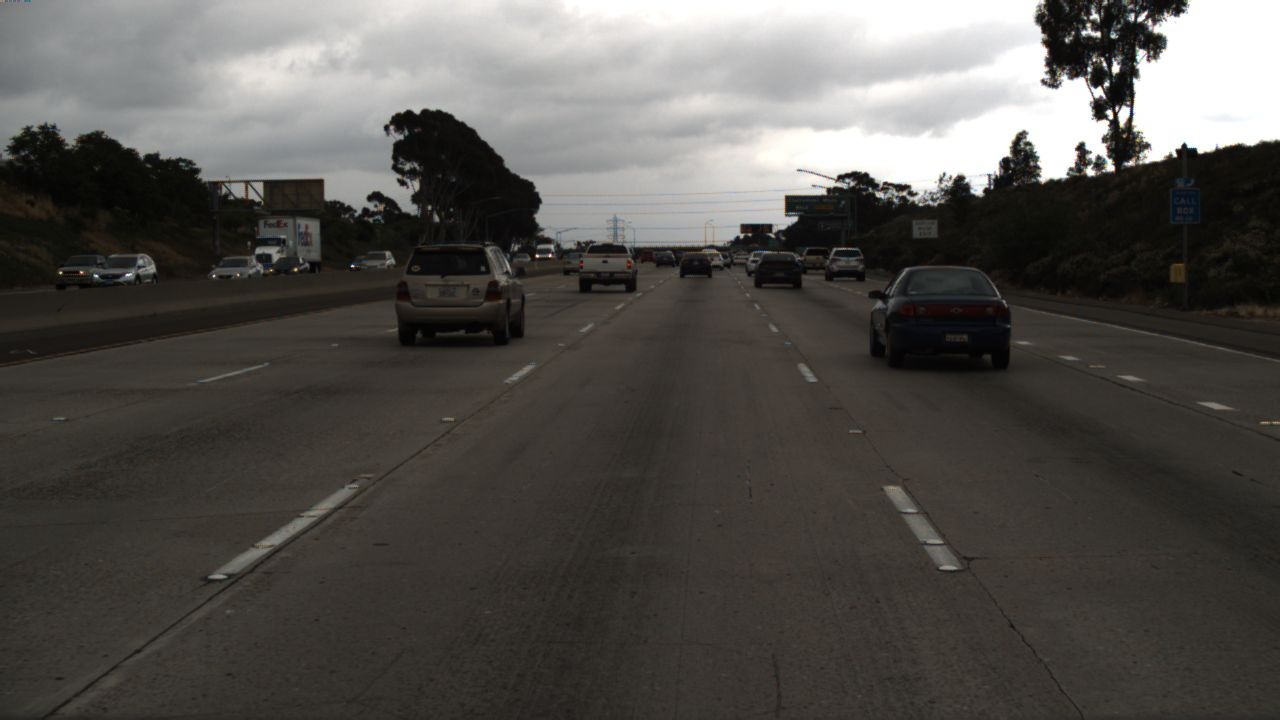}\vspace{3pt}
	\end{minipage}}
	\subfigure[Label]{\begin{minipage}[t]{0.22\linewidth}
			\includegraphics[width=0.7in]{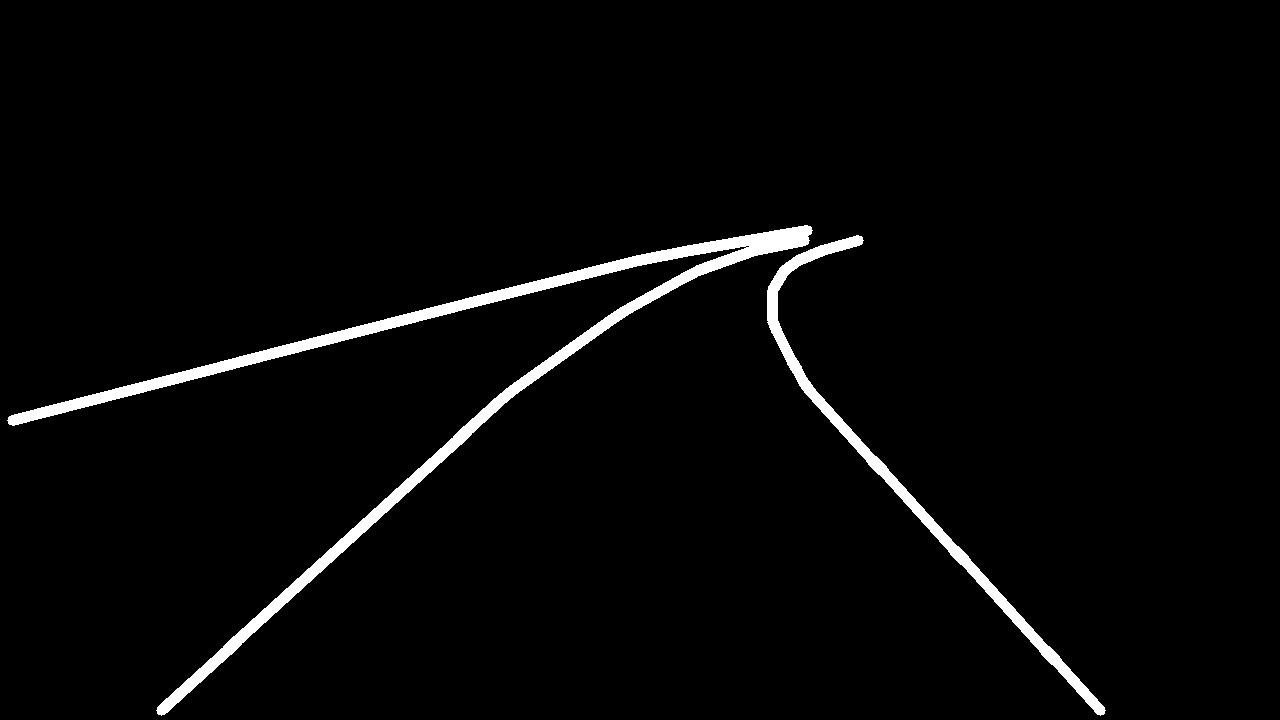}\vspace{3pt}
			\includegraphics[width=0.7in]{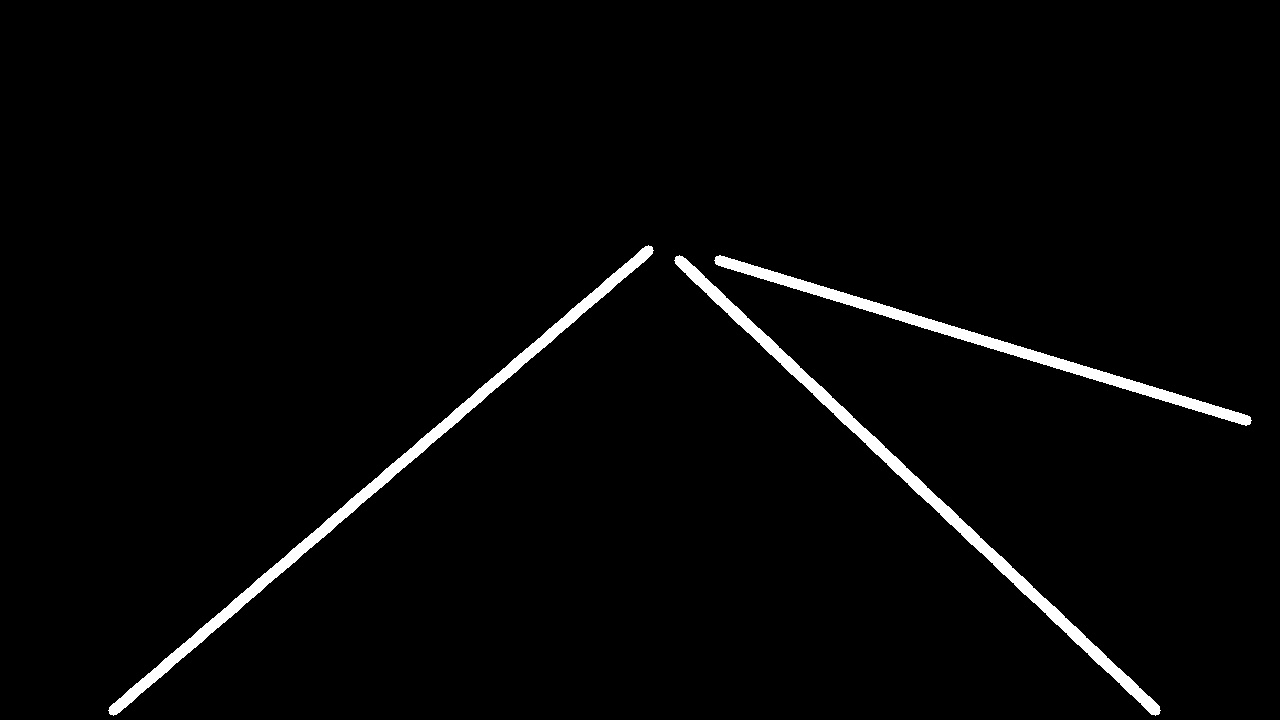}\vspace{3pt}
			\includegraphics[width=0.7in]{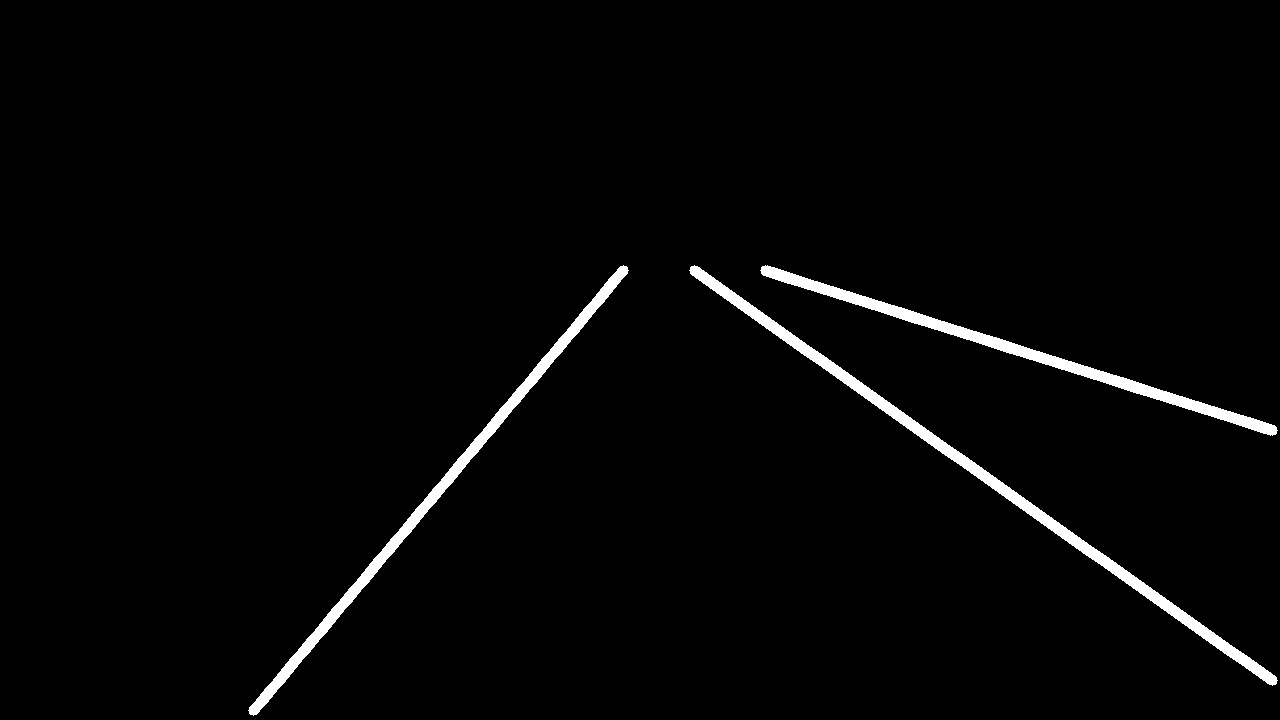}\vspace{3pt}
			\includegraphics[width=0.7in]{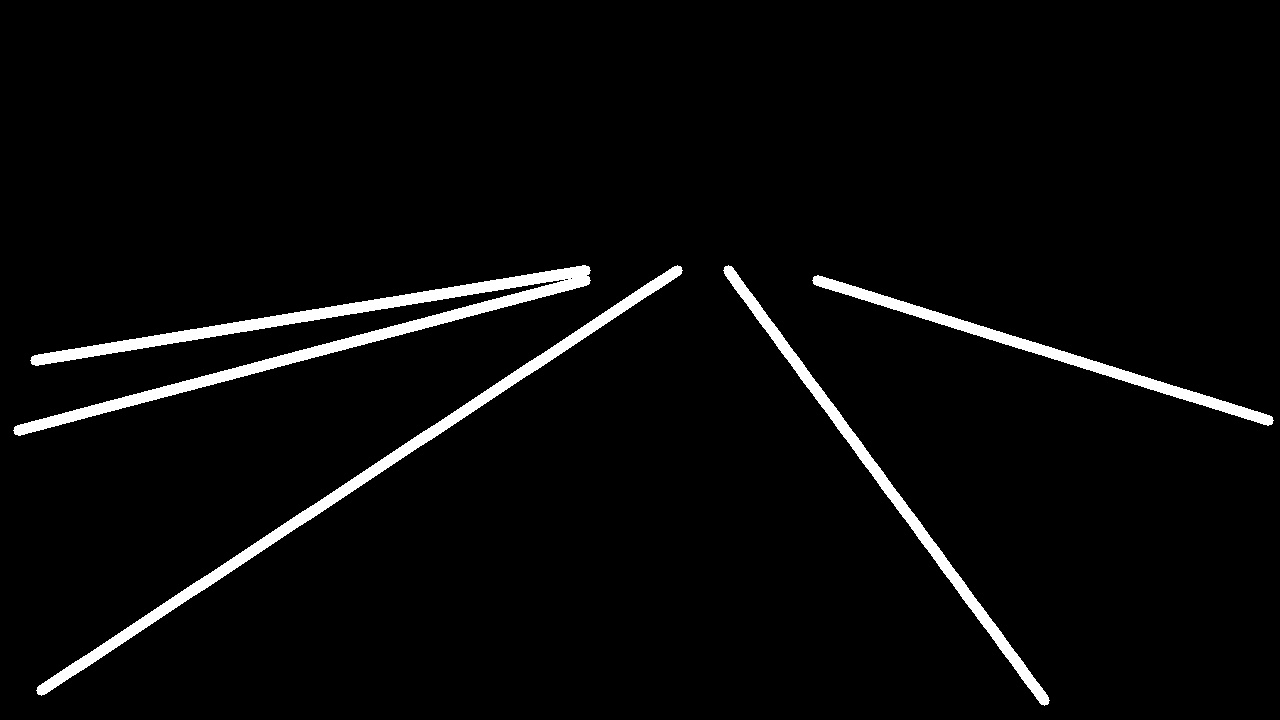}\vspace{3pt}
	\end{minipage}}
	\subfigure[PSPNet]{\begin{minipage}[t]{0.22\linewidth}
			\includegraphics[width=0.7in]{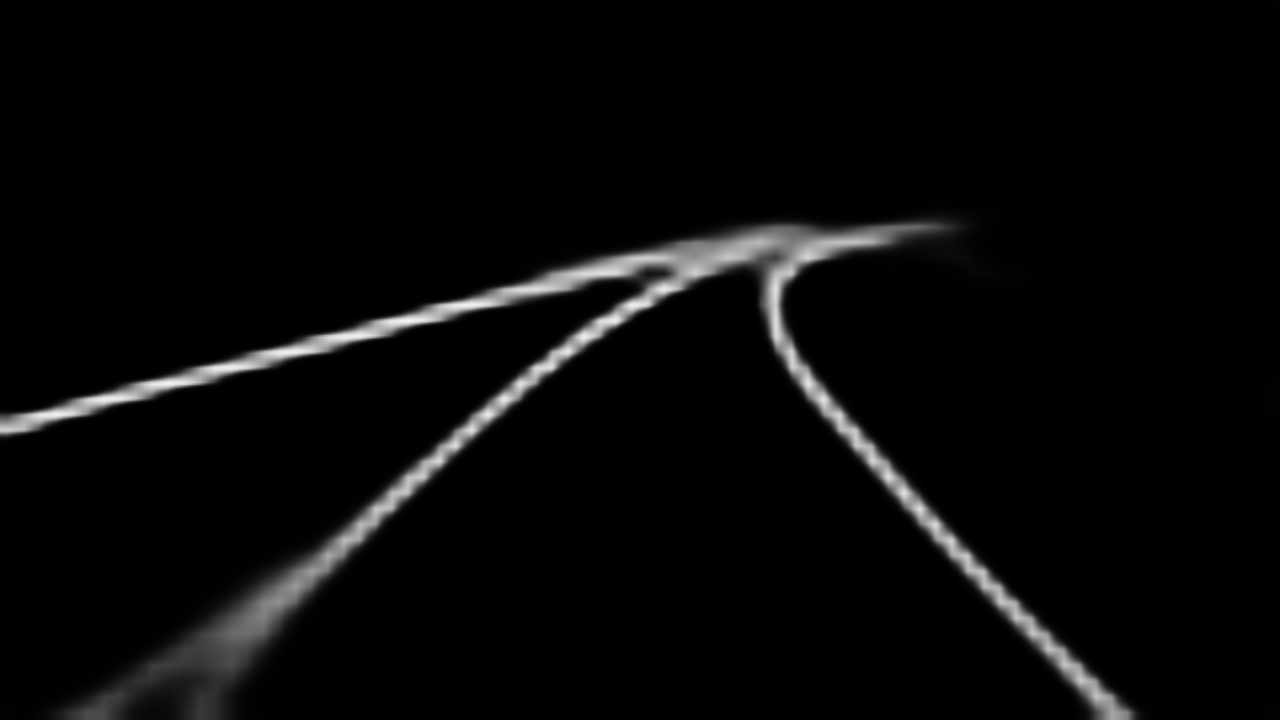}\vspace{3pt}
			\includegraphics[width=0.7in]{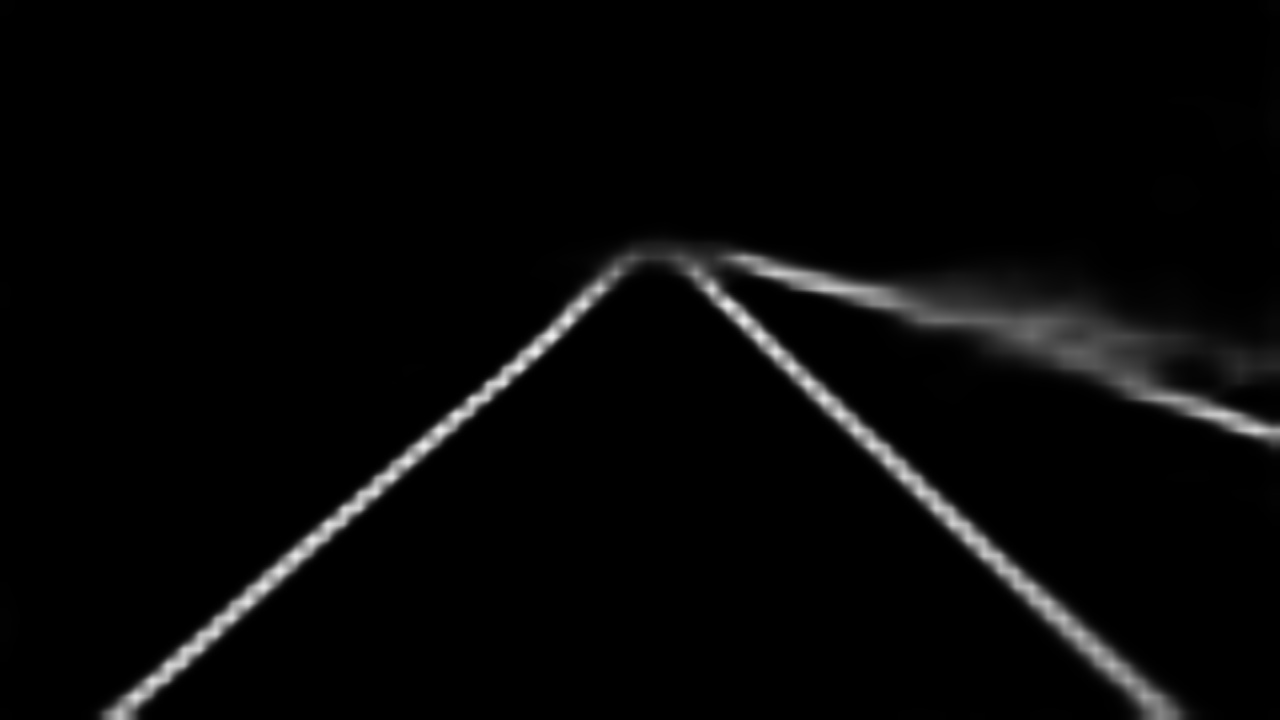}\vspace{3pt}
			\includegraphics[width=0.7in]{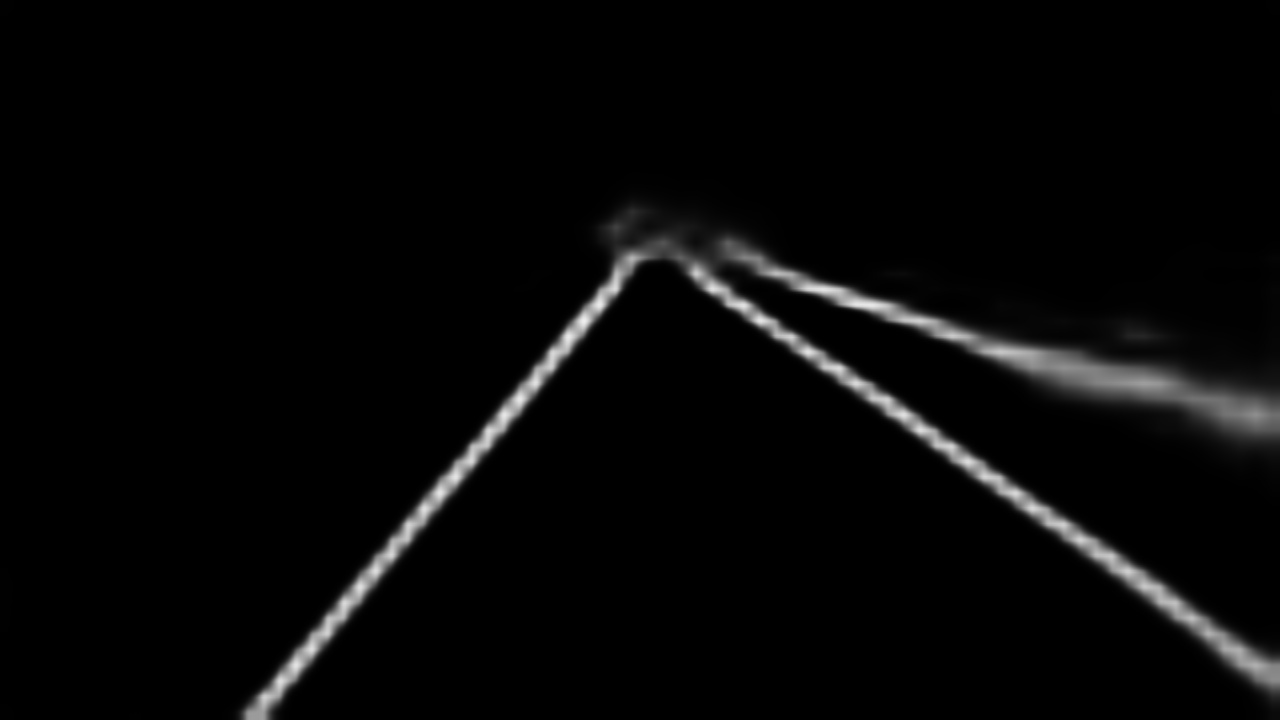}\vspace{3pt}
			\includegraphics[width=0.7in]{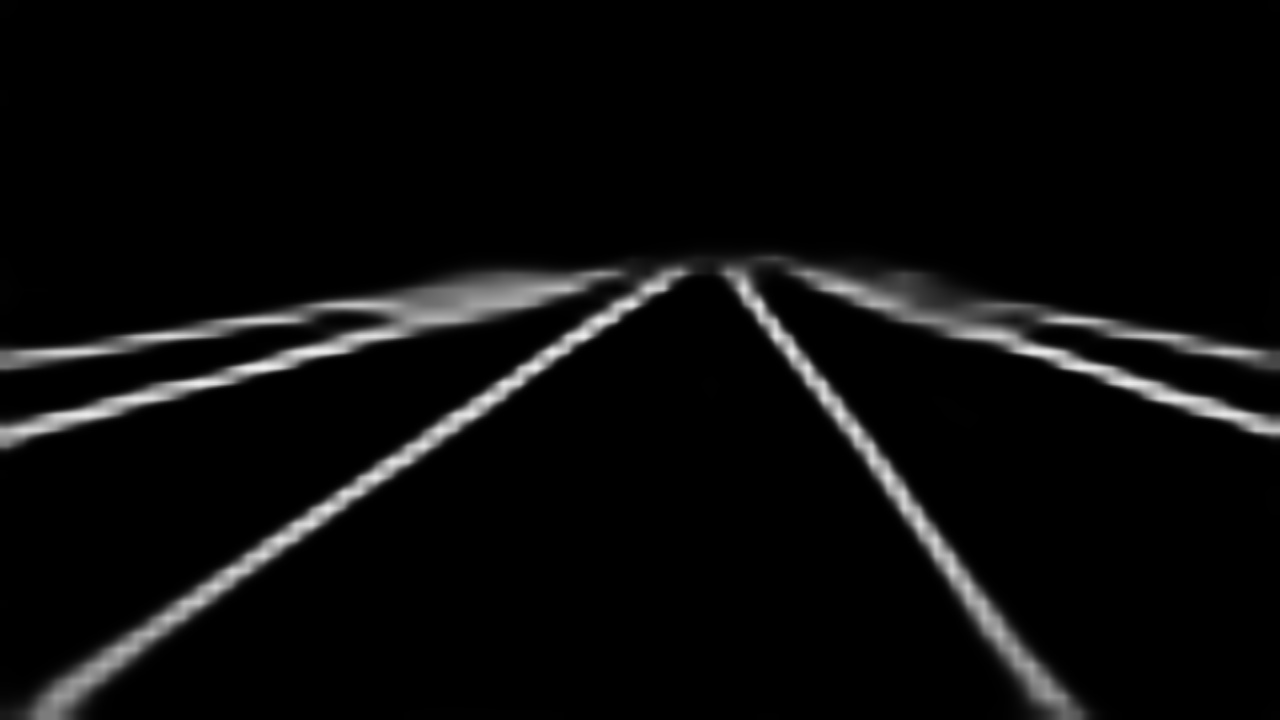}\vspace{3pt}
	\end{minipage}}
	\subfigure[PSP-LGAD]{\begin{minipage}[t]{0.22\linewidth}
			\includegraphics[width=0.7in]{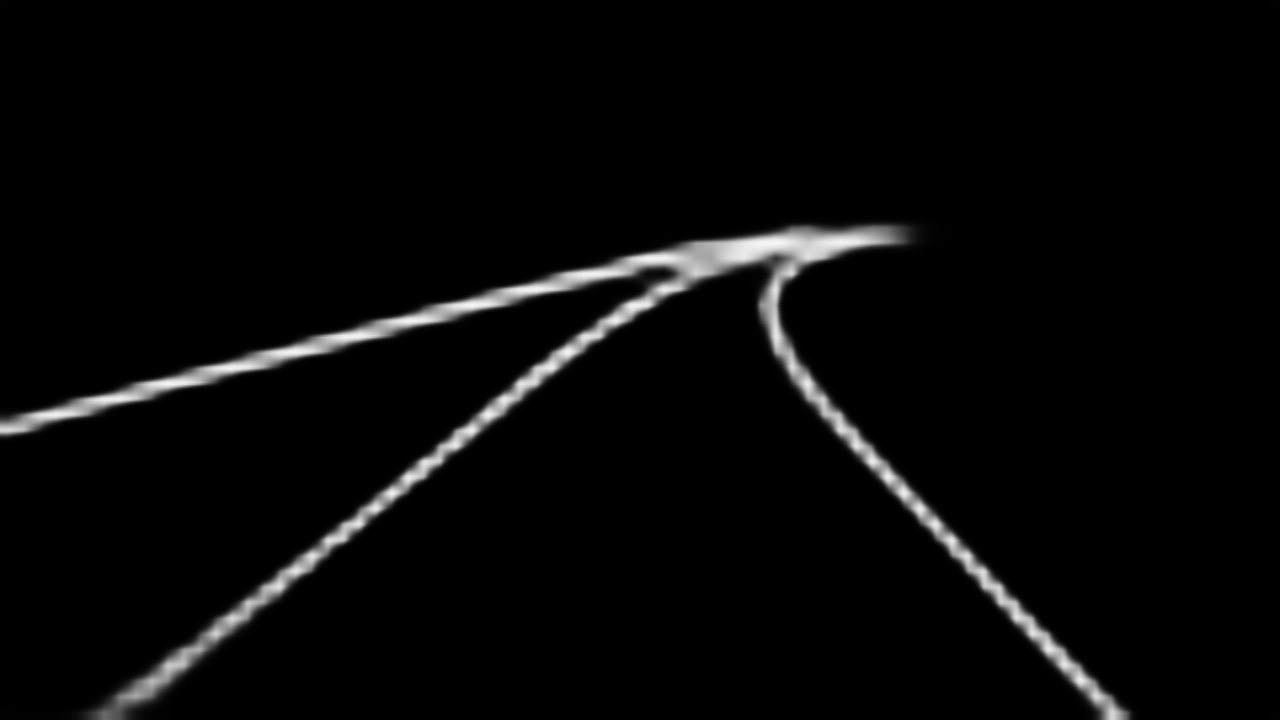}\vspace{3pt}
			\includegraphics[width=0.7in]{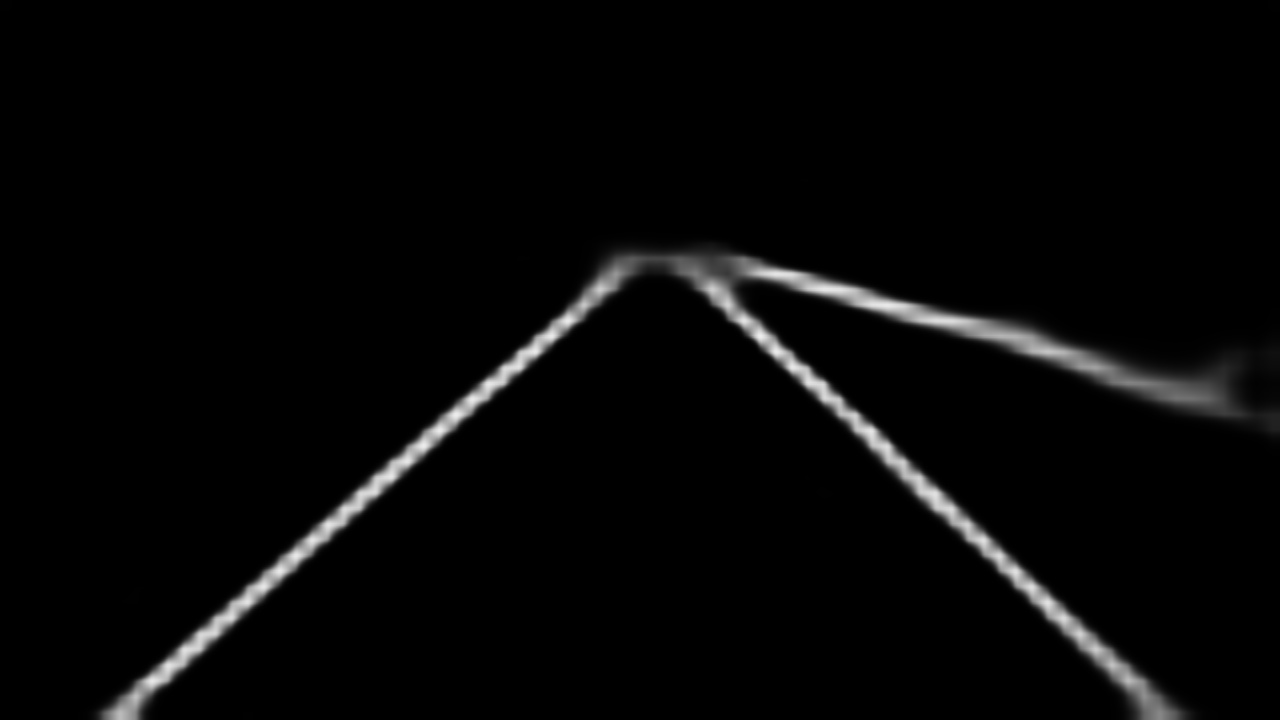}\vspace{3pt}
			\includegraphics[width=0.7in]{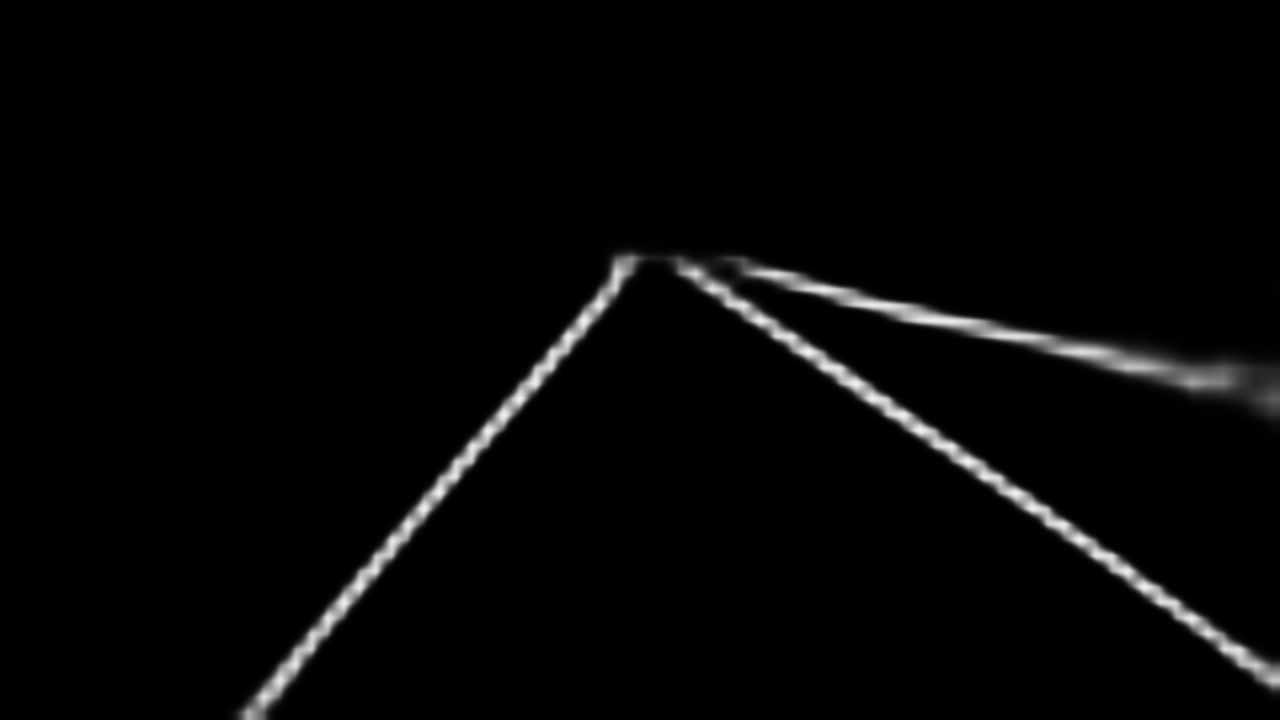}\vspace{3pt}
			\includegraphics[width=0.7in]{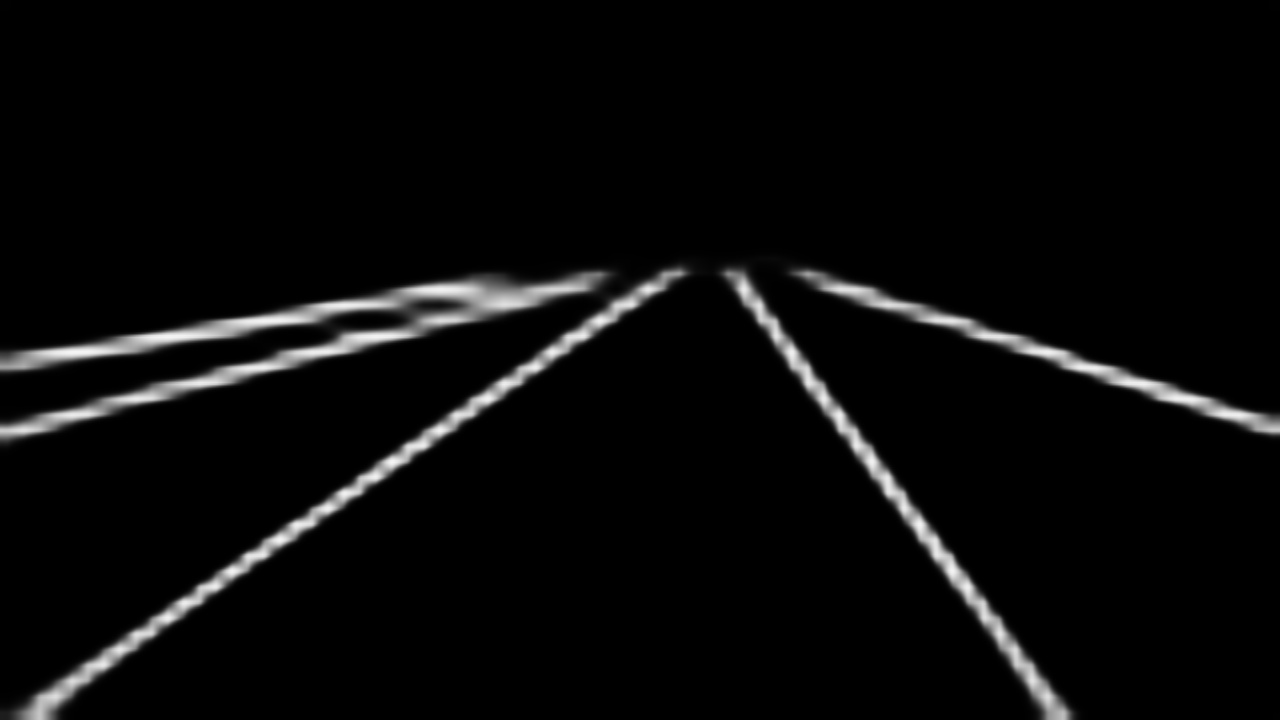}\vspace{3pt}
	\end{minipage}}
	\caption{Visual comparison of PSPNet baseline and PSPNet-LGAD in TuSimple.}
	\label{fig_psp}
\end{figure}

\begin{table}[t]
	\centering
	\caption{Comparison of different models on TuSimple. Three indices are reported, including accuracy, FP, and FN. }
	\begin{tabular}{c|c|c|c}
		\hline
		Model & Accuracy & FP & FN \\
		\hline
		\hline
		ResNet-18 \cite{he2016deep} & 92.69\% & 0.0948 & 0.0822 \\
		ResNet-34 \cite{he2016deep} & 92.84\% & 0.0918 & 0.0796 \\
		BiseNet \cite{yu2018bisenet} & 92.22\% & 0.0884 & 0.0720\\
		PSPNet \cite{zhao2017pyramid} & 93.30\% & 0.0877 & 0.0716 \\
		ENet \cite{paszke2016enet} & 93.37\% & 0.0866 & 0.0705\\
		EL-GAN \cite{ghafoorian2018gan} & 94.9\% & 0.059 & 0.067\\
		LaneNet \cite{neven2018towards} & 96.38\% & 0.0780 & 0.0244\\
		SCNN \cite{pan2018spatial} & 96.53\% & 0.0617 & 0.0180\\
		ENet-SAD \cite{hou2019learning} & 96.64\% & 0.0602 & 0.0205\\
		\hline
		\hline
		\textbf{BiseNet-LGAD (ours)} & 97.48\% & 0.0282 & 0.0103\\
		\textbf{PSPNet-LGAD (ours)} & 97.61\% & 0.0255 & 0.0097\\
		\textbf{ENet-LGAD (ours)} & \textbf{97.65}\% & \textbf{0.0186} & \textbf{0.0065}\\
		\hline	
	\end{tabular}
	\label{tusimple_compare}
\end{table}

\begin{table*}[t]
	\centering
	\caption{Performance ($F_1$ measure) of different methods on CULane. Test results on nine categories and in total are reported. For crossroad, only FP is shown. The second column denotes the proportion of each scenario in the test set.}
	\begin{tabular}{c|c|c|c|c|c|c}
		\hline
		Category & Proportion &SCNN~\cite{pan2018spatial}  &ENet-SAD~\cite{hou2019learning} & \textbf{ENet-LGAD} & \textbf{PSPNet-LGAD} & \textbf{BiseNet-LGAD}\\
		\hline
		\hline
		Normal & 27.7\% & 90.6 & 90.1 & \textbf{91.2} & 90.9 & 91.0\\
		Crowded & 23.4\% & 69.7 & 68.8 & \textbf{68.9} & 68.8 & 68.7\\
		Night & 20.3\% & 66.1 & 66.0 & 66.8 & \textbf{66.9} & 66.7\\
		No line & 11.7\% & 43.4 & 41.6 & \textbf{43.7}& 41.3 & 41.5\\
		Shadow & 2.7\% & \textbf{66.9} & 65.9 & 66.4 & 65.5  & 65.3\\
		Arrow & 2.6\% & 84.1 & 84.0 & \textbf{84.5} & 84.0 & 84.1\\
		Dazzle light & 1.4\% & 58.5 & \textbf{60.2} & 59.9 & 59.8 & 59.8\\
		Curve & 1.2\% & 64.4 & 65.7 & 66.4 & 66.5  & \textbf{66.6}\\
		Crossroad & 9.0\% & 1990 & 1998 & \textbf{1955} & 1973 & 2188\\
		Total & 100.0\% & 71.6 & 70.8 & \textbf{72.0} & 71.5 & 71.4\\
		\hline
		\hline
		Runtime (ms) & - & 133.5 & 13.4 & \textbf{13.4} & 650 & 15.0\\
		Parameter (M) & - & 20.72 & 0.98 & \textbf{0.98} & 250.8 & 5.8\\
		\hline
	\end{tabular}
	\label{culane_compare}
\end{table*}

\begin{figure}[t]
	\centering
	\centerline{\includegraphics[width=9cm]{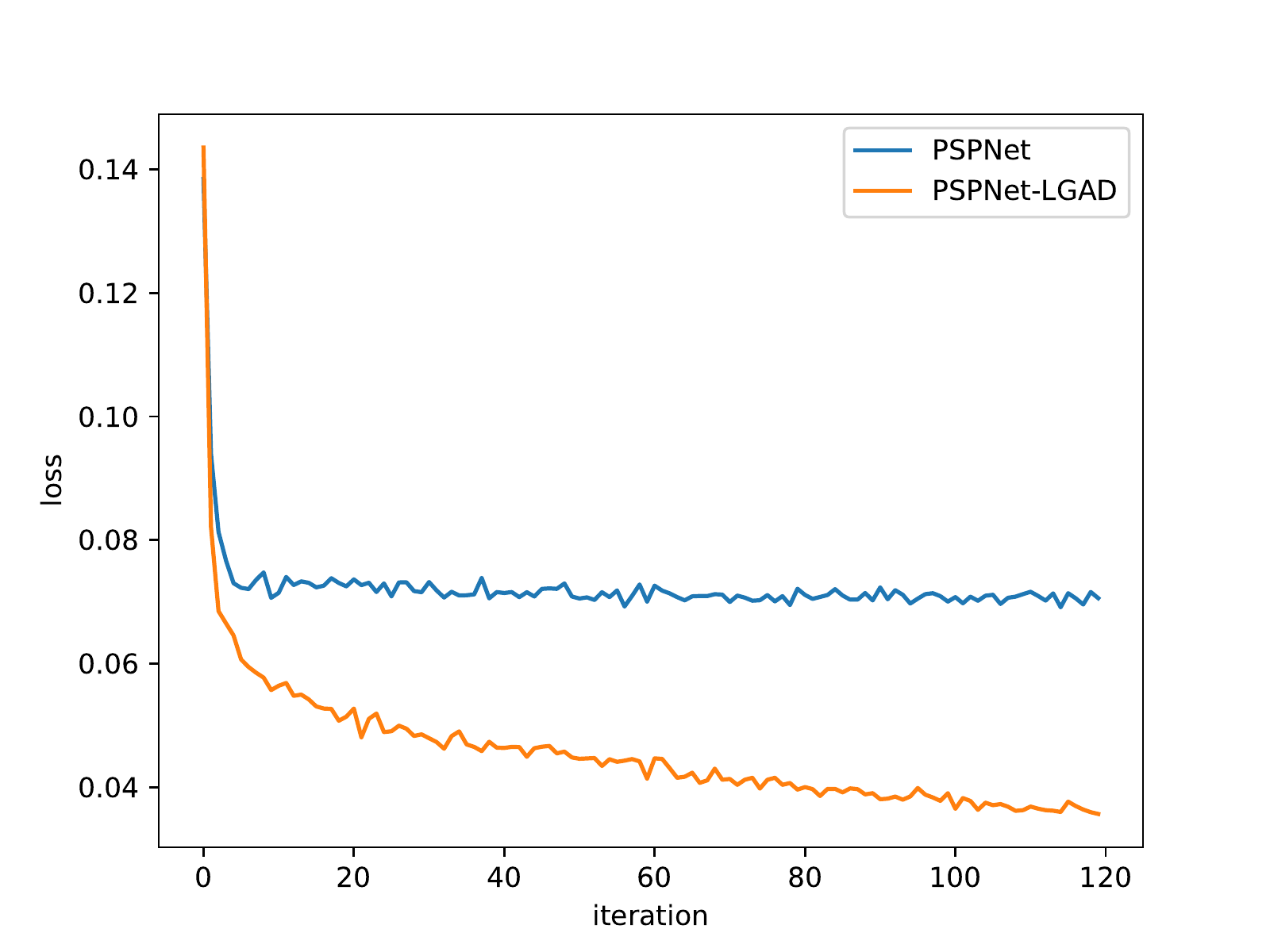}}
	\caption{Comparison of $\mathcal{L}_{seg}(s,\hat s)$ loss curves between PSPNet and PSPNet-LGAD. The loss values are recorded every 10 iterations in the first 4 epochs.}
	\label{loss}
\end{figure}

\subsection{Experiment Results}

Table \ref{tusimple_compare} and Table \ref{culane_compare} summarize the performance of our methods, \ie, BiseNet-LGAD, PSPNet-LGAD, and ENet-LGAD against the baselines and state-of-the-art methods on TuSimple and CULane datasets. Our results are obtained from BiseNet \cite{yu2018bisenet}, PSPNet \cite{zhao2017pyramid} and ENet \cite{paszke2016enet} with distillation in block 2 in context path, block 2 and block 3, respectively. In these experiments, we follow the original distillation optimization strategy, \ie, we make one-way attention transfer between a static pre-trained teacher and a fresh student.

\textbf{Comparison with baselines.} In TuSimple, it is observed that LGAD shows great effectiveness in improving the performance of all three baseline models. LGAD mechanism increases the accuracy of the three baseline models for more than $4\%$, respectively. The qualitative comparison between segmentation results of the baseline PSPNet and PSPNet-LGAD are depicted in Figure \ref{fig_psp}. It can be seen that PSPNet-LGAD generates more precise segmentation results than PSPNet. Besides, more mistakes are made by the baseline model, especially near the vanishing point. To better understand how LGAD works to improve the model performance, we compare the attention maps of the distillation layer between models with and without LGAD in Figure \ref{fig_pspatt}. As can be seen that, when supervised by LGAD, the attention maps of the student network put more weights on task-relevant objects, \eg, lane lines and road curbs. This would significantly improve the segmentation performance.

To further investigate the impact of LGAD, Figure \ref{loss} compares the $\mathcal{L}_{seg}(s,\hat s)$ loss curves between baseline and PSPNet-LGAD. It can be observed that LGAD helps network converges fast as well as avoid converges to a pool local optimum. We think the underline reason is as follows: unlike traditional object segmentation datasets in which the objects are preciously labeled in pixel-level, the lane segmentation datasets only have coarse instance-level lane labels, \ie, each lane is represented by a set of central points. The label imprecision may cause the models hard to converge or easily converge to a pool local optimum. Attention map guided methods such as LGAD proposed in this paper introduce favorable structure limiting conditions and clear converge targets which would benefit network training. 

\textbf{Comparison with other methods.}
On both TuSimple and CULane, ENet-LGAD achieves new state-of-the-art performance. For TuSimple,  ENet-LGAD with an accuracy of 97.65\% outperforms the state-of-the-art ENet-SAD with an accuracy of 96.64\%. Test results on 9 categories of test images of CULane are reported in Tabel \ref{culane_compare}. ENet-LGAD achieves the new state-of-the-art performance with a F-measure of 72.0\% and outperformance other methods on almost all categories. It can be observed that under the supervise of LGAD, model performance is improved effectively on the night and no-line scenarios in which lane lines are not salient.

We also compare the runtime and parameter count of different models in Table \ref{culane_compare}.  Although training the teacher network requires additional computation, the student network requires no extra computation and memory at inference. As can be seen that ENet-LGAD, which outperforms all other methods, has 20 × fewer parameters and runs 10 × faster compared with SCNN. The performance strongly suggests the effectiveness of LGAD.

\begin{table}[t]
	\centering
	\caption{Test results of ENet-LGAD supervised in different layers. Acc denotes the accuracy.}
	\begin{tabular}{c c c c | c}
		\hline
		Layer1 & Layer2 & Layer3 & Layer4 & ACC\\
		\hline
		\checkmark & & & & 97.63\%\\
		 & \checkmark & & & 97.65\%\\
		 & & \checkmark & & 97.63\%\\
		 & & & \checkmark & 97.65\%\\
		& & \checkmark & \checkmark & 97.32\%\\
		\checkmark & \checkmark & & &97.36\% \\
		\checkmark & \checkmark & \checkmark & \checkmark & 97.05\%\\
		\hline
		
	\end{tabular}
	\label{layer}
\end{table}

\begin{table}[t]
	\centering
	\caption{Test Results of ablation studies on TuSimple.}
	\begin{tabular}{l|l}
		\hline
		Model & Accuracy\\
		\hline
		\hline
		ENet  & 93.37 \%\\
		ENet-LGAD & 97.65\% \\
		ENet-CLGAD & 97.62\% \\
		ENet-DS & 95.84\% \\
		ENet-DML & 95.55\%\\
		ENet-FMD & 96.02\%\\
		\hline
		\hline
		BiseNet  & 92.22\% \\
		BiseNet-LGAD & 97.48\% \\
		BiseNet-CLGAD & 97.44\% \\
		BiseNet-DS & 95.89\% \\
		BiseNet-DML & 94.43\% \\
		BiseNet-FMD & 95.13\% \\
		\hline
	\end{tabular}
	\label{ablation}
\end{table}

\subsection{Ablation Studies}
\label{exp_ablation}
\textbf{Distillation Position.} The label-guided distillation can be applied to supervise any layer of the model. We design a group of experiments to test the influence of distillation positions on model performance and the results are shown in Table \ref{layer}. We have a few observations. First, the distillation position is not a decisive factor affecting model performance, \ie, we get more than 3.5\% accuracy boost wherever to place distillation. Second, multi-layer distillation brings no improvement but slightly reduces the test accuracy. The reason may be that single layer distillation is able to help network converge to a desirable state. Too much distillation would slightly harm the feature extraction ability of the deep layers.

\textbf{Optimization strategies.} We compare the optimize strategy of training the teacher and the student collaboratively with training the teacher and the student one by one. The collaborative training strategy results are shown in rows ENet-CLGAD and BiseNet-CLGAD in Table \ref{ablation}, respectively. We can observe that this strategy yields almost equivalent performance compared with ENet-LGAD and BiseNet-LGAD.  The fact indicates the robustness of the proposed attention distillation method. 

\textbf{Comparison with deep supervision.} 
Deep supervision \cite{xie15hed} is a well-known technology to improve model performance without adding inference computation. Here, deep supervision denotes the method that uses the labels directly as supervision for the layers in the network. In our experiment, the supervision is achieved by a $1\times 1$ convolution followed by bilinear upsampling on label maps. The best deep supervision results are shown in rows ENet-DS and BiseNet-DS in Table \ref{ablation}. It can be observed that deep supervision brings improvement though the performance is still worse than LGAD. We also evaluated replace the $1\times 1$ convolution with a $3\times 3$ one but no improvement is brought in this case. 

\textbf{Comparison with deep mutual learning. } 
In the proposed LGAD, the teacher network is trained from label to label. Deep mutual learning (DML) \cite{zhang2018deep} trains two networks with different weight initialization collaboratively and both networks are trained from scene image to label. Similar to LGAD, the two networks have the same structure in DML. The comparison between LGAD and DML is shown in Table~\ref{ablation}. Although both methods bring effective improvement, LGAD outperforms DML. We attribute this to the following reasons. On the one hand, as conventional FCNs can not perform well on lane segmentation task, it is not qualified enough to use them as teachers. On the other hand, the scene image to label trained teacher cannot provide sufficient structure information which is important for segment long-range narrow-width lanes.

\textbf{Comparison with feature map distillation.} We compare the performance of transferring information directly from feature maps with from attention maps, \ie, LGAD. Feature map distillation is added to ENet and BiseNet using $l_2$ loss. We found the improvements over baseline student is minimal (as shown in rows ENet-FMD and BiseNet-FMD in Table~\ref{ablation}). This indicates that the attention maps carry information that is more important to transfer than feature maps.

%------------------------------------------------------------------------
\section{Conclusions}
In this paper, we propose a Label-guided Distillation (LGAD) for lane segmentation. LGAD is a distillation method that uses a teacher network to reinforce the attention maps of the student network. The two networks share the same structure with different inputs, \ie, the lane marks and the target images. LGAD can be easily incorporated in any CNN based networks and do not increase the inference time. LGAD is validated in ENet and achieves consistent performance gain on two standard benchmarks, \ie, TuSimple and CULane. Together, these results indicate that we have discovered a new and generally applicable method for regularizing supervised training of lane segmentation networks. 

\section*{Acknowledgements}
This work was supported by grants from the National Key R \& D Program of China [grant number 52019YFB1600500].

\printcredits

\bibliographystyle{cas-model2-names}

\bibliography{cas-refs}

\end{document}